\pdfoutput=1

\documentclass[11pt]{article}

\usepackage[preprint]{acl}

\usepackage{times}
\usepackage{latexsym}
\usepackage{tabularx}
\usepackage{multirow}
\usepackage{booktabs}
\usepackage{dsfont}
\usepackage{enumitem}

\usepackage{graphicx}
\usepackage{subfigure}
\usepackage{float}
\newtheorem{definition}{Definition}

\usepackage{amsmath}
\usepackage{amssymb}
\usepackage{tcolorbox}
\usepackage{comment}
\usepackage{xcolor}

\usepackage{makecell}
\usepackage{booktabs}
\usepackage{dsfont}
\usepackage[T1]{fontenc}

\usepackage[utf8]{inputenc}

\usepackage{microtype}
\usepackage{enumitem}
\setenumerate[1]{itemsep=0pt,partopsep=0pt,parsep=\parskip,topsep=5pt}
\setitemize[1]{itemsep=0pt,partopsep=0pt,parsep=\parskip,topsep=5pt}
\setdescription{itemsep=0pt,partopsep=0pt,parsep=\parskip,topsep=5pt}
\usepackage{inconsolata}

\newcommand{\argmax}{\mathop{\mathrm{argmax}}\nolimits}
\title{Exploring Forgetting in Large Language Model Pre-Training}

  \author{Chonghua Liao$^1$, Ruobing Xie$^2$  \\
  \textbf{Xingwu Sun}$^2$, 
  \textbf{Haowen Sun}$^1$, 
  \textbf{Zhanhui Kang}$^2$\\
  $^1$ Tsinghua University, 
$^2$ Machine Learning Platform Department, Tencent\\
\texttt{lch22@mails.tsinghua.edu.cn} \quad \texttt{xrbsnowing@163.com}
}
\begin{document}
\maketitle

\newenvironment{tightenumerate}{
\begin{enumerate}[leftmargin=0pt]
  \setlength{\itemsep}{0pt}
  \setlength{\parskip}{0pt}
}{\end{enumerate}}
\newenvironment{tightitemize}{
\begin{itemize}[leftmargin=0pt]
  \setlength{\itemsep}{0pt}
  \setlength{\parskip}{0pt}
}{\end{itemize}}

\begin{abstract}

Catastrophic forgetting remains a formidable obstacle to building an omniscient model in large language models (LLMs). Despite the pioneering research on task-level forgetting in LLM fine-tuning, there is scant focus on forgetting during pre-training. We systematically explored the existence and measurement of forgetting in pre-training, questioning traditional metrics such as perplexity (PPL) and introducing new metrics to better detect entity memory retention. Based on our revised assessment of forgetting metrics, we explored low-cost, straightforward methods to mitigate forgetting during the pre-training phase. Further, we carefully analyzed the learning curves, offering insights into the dynamics of forgetting. Extensive evaluations and analyses on forgetting of pre-training could facilitate future research on LLMs.

\end{abstract}



\section{Introduction}

Catastrophic forgetting~\citep{mccloskey1989catastrophic, ratcliff1990connectionist} poses a significant challenge to the development of models
Traditionally, the challenge of catastrophic forgetting in neural networks is especially pronounced when models are tasked with retaining knowledge across diverse datasets~\citep{sun2020ernie, jin2021lifelong, de2019episodic, wang2020efficient, qin2022elle}.
This issue arises due to the shift in input distribution across different tasks, which leads to the model's inability to remember past knowledge and capability effectively.

Although pioneer efforts have explored the forgetting issue in LLM fine-tuning, which primarily addresses task-specific forgetting, there is a lack of research on finer-grained forgetting in \textbf{pre-training}.
\citet{luo2023empirical},~\citet{wang2023orthogonal}, and~\citet{wu2024llama} focused on forgetting in fine-tuning by measuring the performance of new tasks with continual tuning.
Other efforts~\cite{tirumala2022memorization,biderman2023emergent} studied sample-level memorization, where some experiments imply the existence of forgetting in LLM pre-training.
Nonetheless, these studies have devoted limited attention to systematically exploring and quantifying the forgetting in pre-training.

Forgetting in pre-training is a critical issue that must be addressed. It is prevalent among current LLMs and significantly affects their performance. Usually, models are believed to acquire various factual knowledge during the pre-training phase, and during the fine-tuning phase, they enhance their task-related capabilities~\citep{chang2024large}. Intuitively, LLMs may give unsatisfactory replies to fact-relevant queries, even when the necessary information was present in the pre-training data. This indicates forgetting. Despite being easily noticed, measuring this forgetting in pre-training is very challenging. In contrast to works studying fine-tuning that measure with specific task-related metrics (e.g., QA accuracy), the pre-training data is too diverse, inherently consisting of dozens of tasks, making it almost impossible to use a specific ability-related metric to reflect forgetting. Moreover, there's almost no metrics designed for forgetting. General metrics such as perplexity (PPL) are also shown to be insensitive in measuring forgetting in pre-training~\citep{gupta2023continual}.
This raises a pertinent question: (1) \textit{How to correctly recognize and quantify forgetting in pre-training?}

After correctly understanding and assessing the phenomenon of forgetting, which we address by introducing innovative metrics, we then shift our focus to exploring \textit{lightweight} methods aimed at mitigating this issue. Inspired by the proven success of memory replay methods in combating forgetting \textbf{during dataset shifts}, as shown in~\citep{de2019episodic, wang2020efficient}, we delve into the inquiry: (2) \textit{Can these methods also mitigate forgetting during the pre-training phase?} 

Then, following the above investigation, we proceed to examine the interplay between memory replay and the learning dynamics. That is, we emphasis on elucidating the models' forgetting curves. Inspired by the human learning premise that a higher review intensity can decelerate the forgetting rate~\citep{loftus1985evaluating}, we aim to observe whether the aspects of knowledge replay and learning intensity in models exhibit similar phenomena on the learning curve as those inspired by human learning processes. This observation could, in turn, guide the design of memory replay methods. With this in mind, we pose the inquiries: (3) \textit{Do models display forgetting patterns akin to human learning? Can these patterns guide the design of memory replay to further mitigate forgetting?}



To address the above questions, we conducted a series of explorations that progressively and deeply advance in logic.
We first magnify the forgetting issue by building a didactic scenario, and scrutinize the limitation of conventional metrics (e.g., PPL) in identifying forgetting. 
Next, we focus on \textbf{the recall ability of entity-related information}, one of the most explicit and significant indicator of forgetting during pre-training. 
We propose four novel entity-related metrics and experimentally confirm the existence of forgetting during pre-training. Within a standard pre-training setting, we present several simple and lightweight memory replay strategies, and show that simple and cost-effective replay strategies can effectively mitigate forgetting. Finally, drawing an analogy to the human memory curve, we examine how the metrics of recently learned samples evolve over the course of further learning. We then explore the impact of short-term, high-frequency learning on the model's memory retention, shedding light on future pre-training designs aimed at mitigating forgetting.



Our main contributions are: (1) We systematically explore and quantify the phenomenon of pre-training forgetting through new entity-focused metrics. (2) We examine the effectiveness of memory replay in reducing pre-training forgetting. (3) We further examine how short-term, high-frequency learning affects the forgetting curve.



\section{Related Work}


\paragraph{Catastrophic Forgetting in Language Models.} 
Neural networks often experience catastrophic forgetting when changing data distribution~\citep{mccloskey1989catastrophic, ratcliff1990connectionist}. Various strategies have been proposed, such as simultaneous training of new and old tasks~\citep{sun2020ernie}, incremental lifelong pre-training~\citep{jin2021lifelong}, and the incorporation of episodic memory~\citep{de2019episodic}. Other approaches include meta-lifelong frameworks~\citep{wang2020efficient} and function-preserved model expansion~\citep{qin2022elle}. However, most of these studies do not explore single data distribution scenarios. Our study uniquely focuses the pre-training phase, offering fresh insights into forgetting.

\paragraph{Example Forgetting and Forgetting During Pre-training.}  Despite significant research on forgetting, there is limited investigation within the context of a single task.~\citet{toneva2018empirical} first defined example forgetting.~\citet{tirumala2022memorization} explored forgetting dynamics in LLMs.~\citet{biderman2023emergent} studied model behavior forecasting, while~\citet{gupta2023continual} examined warm-up strategies in continual pre-training. However, a detailed formalization and quantification of forgetting during pre-training using metrics has been lacking—this is where our research steps in.

\section{Existence of Pre-training Forgetting}

\subsection{Intuition on Pre-training Forgetting}
\label{2.1}

First, to test if there is a forgetting trend, we explore whether, \textbf{after pre-trained}, an LLM \textit{exhibits a pattern of decreased performance on earlier seen samples}.
To test this, a direct approach is: after training, we obtain a checkpoint and then \textbf{use this exact checkpoint} to test on samples in the sequence they were encountered during training. This helps us to assess the model's retention of information over time. We aim to assess if existing metrics like PPL can monitor trends throughout training.

\subsubsection{Setup and PPL}
We uniformly sampled a subset with 4.9e8 tokens from SlimPajama~\citep{cerebras2023slimpajama}. Then we conducted standard and memory-replay pre-training. \textit{To reflect the model's training progression}, a test set was created by sequentially segmenting the training data according to the training steps and uniformly sampling 1/100 of each segment. PPL is plotted against the number of training tokens processed, with the test set's token count scaled to match the model's exposure. More details are in Appendix \ref{appendix metric part 1.1}.

\begin{figure}
    \centering
    \includegraphics[width=0.75\linewidth]{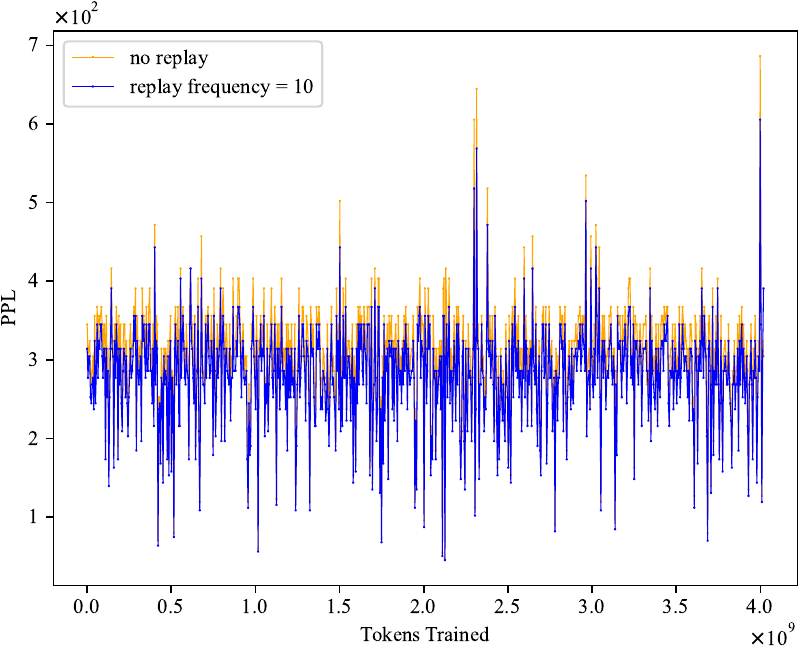}
    \caption{Perplexity (PPL) of the GPT-2 XL model on uniformly sampled 1/100 segments of the training data. Considering forgetting does help the performance.}
    \label{fig:SlimPajama_0.5B}
\end{figure}

\noindent
\textbf{Results:}
The result is shown in Figure~\ref{fig:SlimPajama_0.5B}. This indicate that: (1) The model shows stable performance across early and late training data, with comparable PPL, challenging the hypothesis of higher early training perplexity. This suggests either that forgetting is not occurring, contrary to our understanding, or that forgetting exists but is not captured by PPL. (2) Model with replay during pre-training shows better performance, with a notable drop in average PPL (280.66 with replay vs. 303.63 without), \textit{indirectly confirming the existence of forgetting} through performance gains from repeated learning.





\subsection{The Failure of Traditional Metrics}
\label{2.2}
In previous experiments, we realized that detecting forgetting was challenging in a single pre-training dataset due to the \emph{uniformity of the data}. To tackle this, we build an A+B dual-dataset scenario, aiming for datasets A and B to be similar yet slightly different to magnify forgetting effects. With dataset A being much smaller than B, we aim to avoid overfitting on it. This emulates the scenario in an actual single pre-training dataset where A represents a little portion of the early data at risk of being forgotten as training advances with an ever-growing pool of data. Beyond practical convenience, this is also a common setting for continuing pre-training.
 

\noindent
\textbf{Setup:}
We uniformly sample a subset from dataset A as a test set and then train on dataset B, evaluating the model to observe forgetting of dataset A. We conduct two experiments, employing the OpenWebText~\citep{Gokaslan2019OpenWeb} dataset ($\sim$8B tokens) for dataset A in one experiment, and a uniformly sampled subset from the Pile~\citep{gao2020pile} ($\sim$ 13B) for the other. Dataset B is constituted by a uniformly sampled subset ($\sim$ 49 B) tokens from SlimPajama. More details are in Appendix \ref{appendix metric part 1.2}.
%
Our investigation into forgetting in pre-training, while pioneering, is bounded by computational limitations. The requirements in the following sections, estimated at $\sim$10,000 GPU hours on 8 NVIDIA A100 GPUs (40 GiB VRAM), present a significant challenge. This indicates that utilizing a 1.5B model to complete all subsequent experiments would require 30,000 GPU hours ($\sim$150 days). Such computational costs are prohibitive for a research exploration. To allocate more computational resources towards exploration of phenomena across dozens of experiments and to gain a deeper understanding, we decided to conduct all subsequent experiments on GPT-2.

\begin{figure}[!ht]
\centering
    \subfigure[PPL on OpenWebText]{%
      \includegraphics[width=0.45\linewidth]{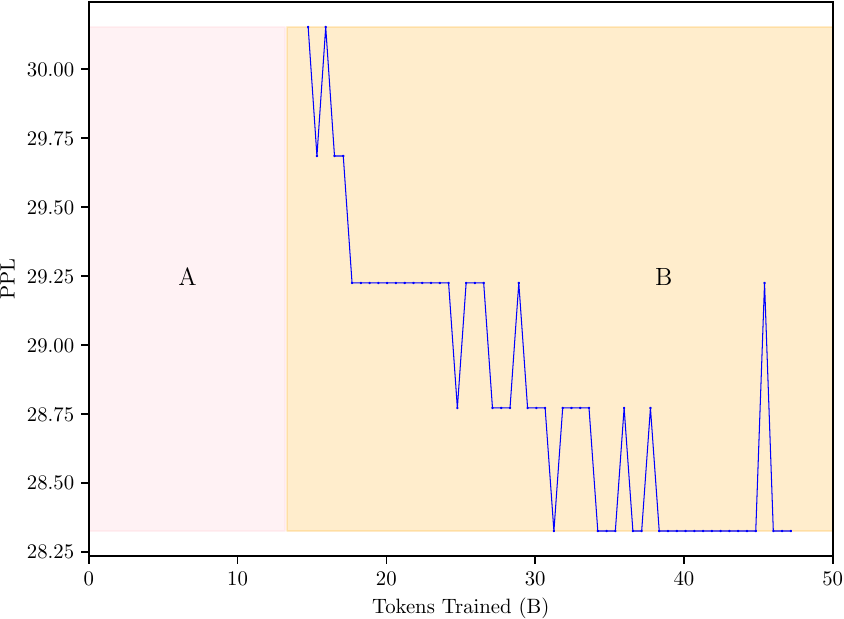}
    }
    \subfigure[PPL on the Pile]{%
      \includegraphics[width=0.45\linewidth]{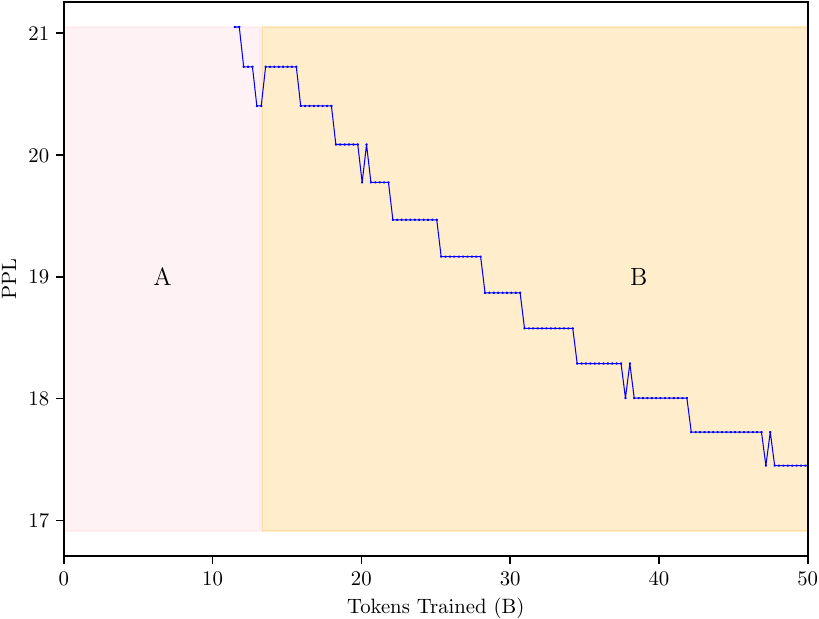}
    }\
    \subfigure[M(f) on the Pile]{%
      \includegraphics[width=0.45\linewidth]{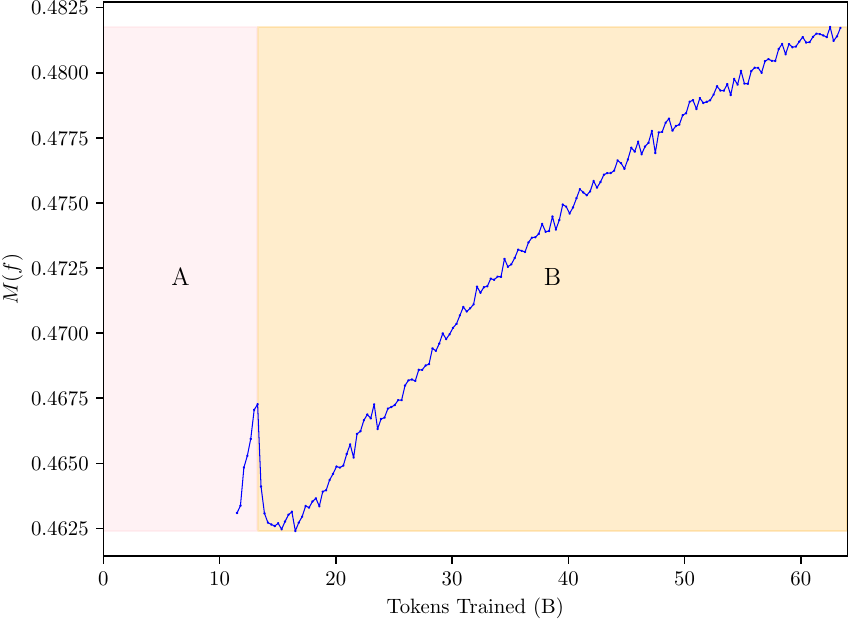}
    }\
    \caption{(a), (b): PPL of the eval of dataset A in relation to the number of trained tokens. A is a subset of OpenWebText(a) or the Pile(b). The fluctuating PPL is not a good indicator of forgetting. (c): M(f) of the eval for the Pile. At the A-to-B dataset transition, M(f) shows negligible changes, where we capture the subtle  signal of forgetting, and then consistently increases.}
    \label{fig:AB_PPL}
\end{figure}

\vspace{0.1pt}
\noindent
\textbf{Results of PPL:} The results in Figure \ref{fig:AB_PPL} (a)(b) reveal an unexpected trend: contrary to expectations of increasing PPL for dataset A as a sign of forgetting during dataset B's training, the PPL for dataset A actually decreased in both setups. Even during the transition between datasets, PPL showed minimal signs of forgetting.


\subsubsection{M(f) Metric}
Recognizing the shortcomings of PPL in accurately measuring forgetting, we turned to the M(f) metric introduced by~\citet{tirumala2022memorization} for evaluation.
The detailed definition of M(f) is:
\begin{definition}
\label{def:memorization}
Let $V$ denotes the vocabulary size. The set $C$ consists of contexts $(s, y)$, $s$ is an incomplete text and $y$ is the correct token index. $f: S \to \mathbb{R}^V$ is a language model. A context $c$ is memorized if $f(s)$'s maximum value corresponds to $y$, i.e., $\argmax_{\mathbf{w} \in \mathbb{R}^V} f(s) = y$. We assess the fraction of contexts memorized using $M(f) = \frac{\sum_{(s, y) \in C} \mathds{1} \{\argmax(f(s))= y \}} {|C|}.$
\end{definition}


\noindent
\textbf{Results of M(f):}
In this experiment, we continued to employ the A (the Pile) + B (SlimPajama) setup and evaluated the model throughout the entire training process. We also continue to use a uniformly sampled 1/1000 part of A as the test set. We observed that at the transition from dataset A to dataset B, M(f) exhibited subtle fluctuations. Subsequently, as training progressed on B, the test set's performance, demonstrated a continuous improvement. The results are given in Figure \ref{fig:AB_PPL}.



It is plausible to hypothesize that PPL's probabilistic averaging inherent may not accurately reflect forgetting for common tokens due to their high prediction accuracy, potentially masking information loss for less frequent elements. In contrast, the M(f) metric's binary evaluation is more sensitive to memory errors, offering a clearer view of the model's retention of critical information, essential for understanding catastrophic forgetting.

\subsubsection{Limitation Leads to Underestimate}

Certainly, it is important to acknowledge that both metrics have limitations in capturing forgetting. Our observations indicate that throughout the training process, after the model completed training on dataset A and transitions to dataset B, both metrics show a continuous improvement, with subtle signs of forgetting at the transition point. This suggests a plausible hypothesis: The metrics' inability to account for the token difficulty lead to an underestimation of forgetting, as they are \textbf{dominated by features that are inherently resistant to forgetting}, such as common tokens and simple, everyday text. These features may not exhibit significant prediction errors when the dataset changes, thereby obscuring the true extent of the model's forgetting.

\begin{tcolorbox}[leftrule=1.0mm,top=0.mm,bottom=0.0mm]
\textbf{Takeaway 1:} PPL and M(f) metrics potentially mask true forgetting, as their bias towards easy-to-remember elements can underestimate the model's memory decline across dataset shifts.
\end{tcolorbox}

\section{New Entity-related Metrics for Measuring Pre-training Forgetting}



\subsection{How to Understand Pre-training Forgetting}

Building upon the findings presented, a pertinent inquiry emerges: Which segments of the dataset should be scrutinized to gain a comprehensive understanding of the forgetting phenomenon?

We argue that during pre-training, the focus should be on the forgetting associated with \textbf{entity-related information}. We posit that the capabilities imparted to a model by a dataset can be broadly categorized into two components: information related to entities and task-specific competencies. 
(1)  As demonstrated by~\citet{sorscher2022beyond}, the power law scaling of error shows that many training examples are redundant, and in data-rich scenarios, pruning should focus on retaining challenging examples. Entity-related information, which is less frequent~\citep{penedo2023refinedweb}, is crucial for users' perception of forgetting in LLMs, as it's harder to determine if the loss of abstract capabilities is due to model limitations or forgetting, making entity information key in pre-training.
(2) We also considered the approach of Supervised Fine-Tuning (SFT), which involves training on instructional data. This phase of training enhances the model's capabilities for downstream tasks, and we view it as a stage where the emphasis is on augmenting the model's competencies. Nevertheless, for the pre-training phase, our focus is more directed towards the acquisition of entity information.
(3) Comparing with the forgetting of entities, the forgetting of other content, such as capabilities related to downstream tasks, is more challenging to define and remains ambiguous. Entities serve as an optimal vehicle for exploring the phenomenon of forgetting within our cognitive framework.

\begin{figure}[h]
\centering
    \subfigure[$M_{\text{ex}}$]{%
      \includegraphics[width=0.47\linewidth]{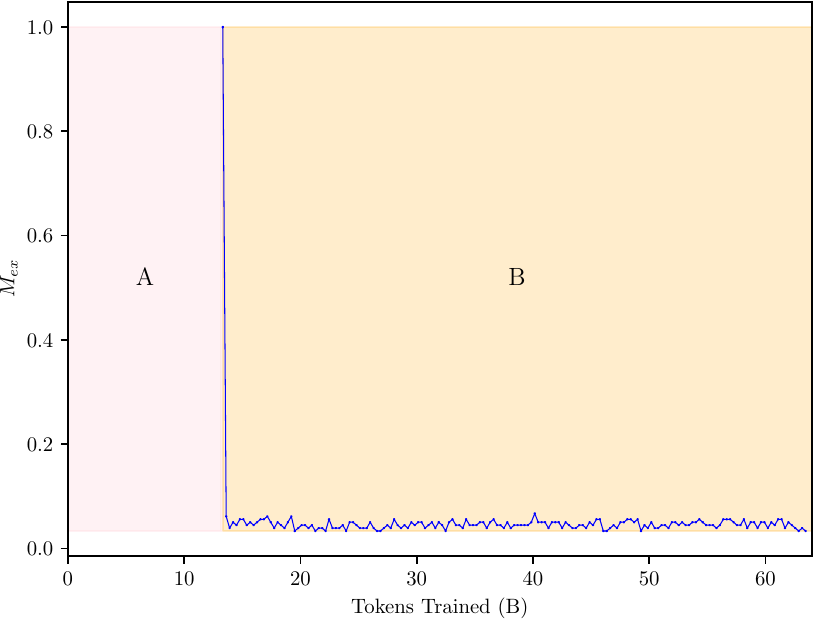}
    }
    \subfigure[$M_{\text{in}}$]{%
      \includegraphics[width=0.47\linewidth]{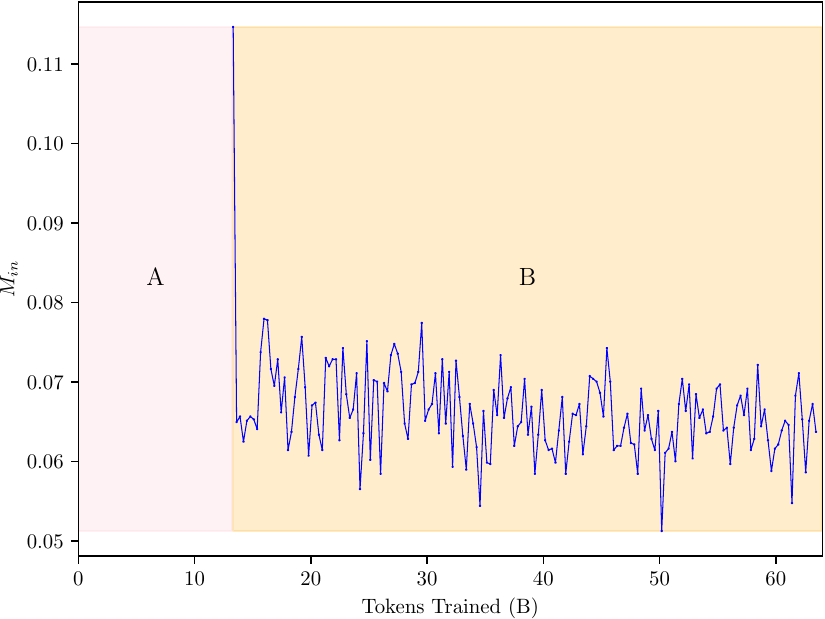}
    }\
    \subfigure[PPL$_{\text{ent}}$]{%
      \includegraphics[width=0.47\linewidth]{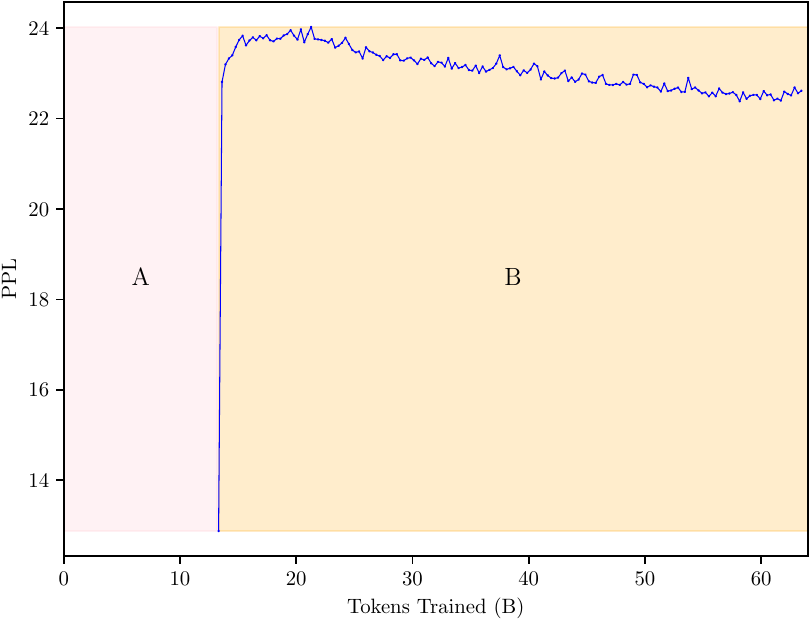}
    }
    \subfigure[M(f)$_{\text{ent}}$]{%
      \includegraphics[width=0.47\linewidth]{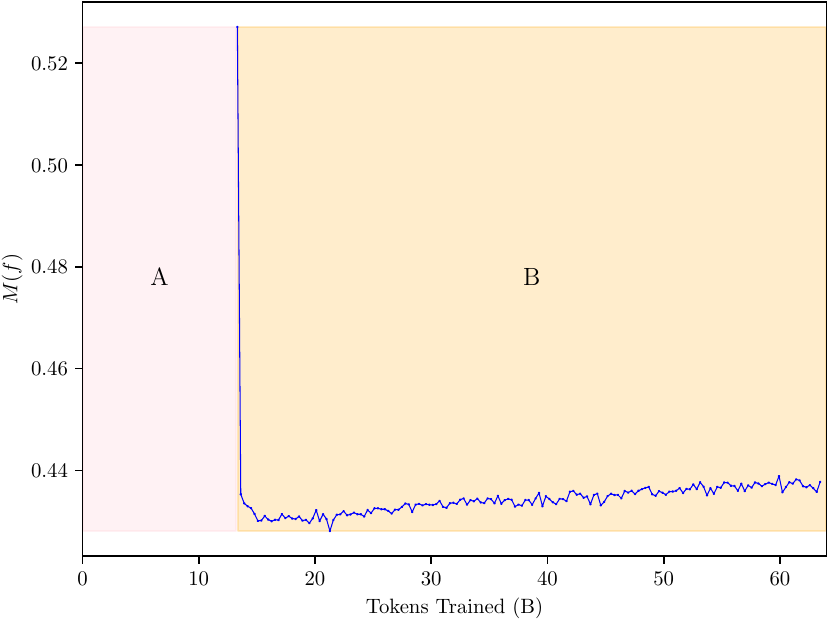}
    }
    \caption{Training dynamics (A (Pile) $\rightarrow$ B (SlimPajama)): entity-focused evaluation set from A reveals marked metric degradation during the A-to-B transition. Besides, traditional metrics on entity-focused samples such as PPL$_{\text{ent}}$ and M(f)$_{\text{ent}}$ exhibit partial recovery during training B. This implies that even for entity-related samples, conventional metrics still focus on information that is less related to entities, which can continue to improve with further learning. }
    \label{fig:result}
\end{figure}
\subsection{Our Proposed Entity-related Metrics}
\label{new metric}

To evaluate forgetting of entities, we follow the memorization score~\citep{biderman2023emergent} and introduce new metrics. These new metrics resemble entity-focused question answering. For further elaboration on the design and illustrative examples of our metrics, please refer to Appendix \ref{appendix metric part 2}.

(1) \textbf{$\bf{M}_{\text{in}}$}: 
Intuitively, this evaluates the model's capacity to \textit{output entity-related details given its context}. We select all samples \( S \) containing a set of entities \( C \). For each sample \( \mathbf{s_i} \in S \), we locate the entities and use the 32 preceding tokens as input, ensuring the entity \( \mathbf{c_j} \in C \) is at the end. Given \( \mathbf{s_i} \), we then greedily decode 32 tokens \( \hat{\mathbf{o}}=(o_1, o_2, ..., o_{32}) \). The original next 32 tokens of $s_i$ \( (t_1, t_2, ..., t_{32}) \) is our target output. The accuracy is defined as \( M_{\text{in}} = \frac{\sum_{\mathbf{s_j} \in S} \sum_{i=1}^{32} \mathds{1} \{o_i = t_i\}}{32|S|} \).

(2) \textbf{$\bf{M}_{\text{ex}}$}: Intuitively, this tests if the model can \textit{recall an entity from the context where the entity is implied but not directly mentioned}. Similar to $M_{\text{in}}$, for a sample $\mathbf{s_i}$ containing entity $\mathbf{c_j}$, we use the preceding 32 tokens as input (excluding $\mathbf{c_j}$). After greedy decoding of 32 tokens $\hat{\mathbf{o}}$, we calculate $M_{\text{ex}}=\frac{\sum_{\mathbf{s_i} \in S}{\text{is\_substring}(\mathbf{c_j}, \mathbf{\hat{o}})}}{|S|}$, where $\text{is\_substring}(\mathbf{a}_1, \mathbf{a}_2)$ returns 1 if $\mathbf{a}_1$ is a substring of $\mathbf{a}_2$ and 0 otherwise.

Besides, we also adopt two entity-centric metrics \textbf{PPL$_{\text{ent}}$} and \textbf{M(f)$_{\text{ent}}$}, which measure existing metrics PPL and M(f) on entity-involved samples.

\noindent
\textbf{Setup:}
We continue to leverage the A+B dataset configuration to accentuate the phenomenon of forgetting, employing the A (the Pile) + B (SlimPajama) dataset setup and training the model on both datasets. Given that A and B are commonly used general-purpose datasets with similar sources, they exhibit no significant differences in text style. Testing is conducted during the training of dataset B. 

We proceed by analyzing frequencies, identifying a set of entities more frequently found in A but less found in B. Using this set, we curated an test set from A and monitored its metrics during B's training to measure the forgetting effect due to less exposure in B. See Appendix \ref{appendix metric part 2} for more details.

\noindent
\textbf{Results:}
In Figure \ref{fig:result}, we have demonstrated the following: (1) When evaluating forgetting on entity-related data, a significantly more pronounced decline is noted, with a notably slow recovery of metrics even during continued training. (2) In evaluations focusing on a subset of data that is rich in samples from source A compared to B, traditional metrics like PPL and M(f) show a recovery. This apparent recovery may be due to less forgettable elements in the data. (3) Comparatively, the newly proposed metrics \( M_{\text{ex}} \) and \( M_{\text{in}} \) exhibit a more difficult recovery, which aligns closely with our expected manifestation of forgetting. This makes them more suitable for indicating forgetting.

\begin{tcolorbox}[leftrule=1.0mm,top=0.mm,bottom=0.0mm]
\textbf{Takeaway 2:} Our newly proposed entity-related metrics, $M_{\text{ex}}$ and $M_{\text{in}}$, exhibit a more noticeable decline and difficult rebound, offering a clearer reflection of the forgetting phenomenon.
\end{tcolorbox}
\section{Explorations on Memory Replay}
With the introduction of our new entity-related metrics, we proceed to an intuitive exploration, specifically investigating whether simple and lightweight design approaches can alleviate this phenomenon. Inspired by~\citet{de2019episodic}, we introduce novel methods for episodic memory replay. We incorporate a module that retains a record of examples. During the learning period, we periodically draw a uniform sample from the memory's stored examples to conduct gradient updates. 

Although other types of methods to reduce \textit{task-level} forgetting during fine-tuning exist, like BERT-based memory~\citep{de2019episodic} and function-preserved expansion~\citep{qin2022elle}, they are computationally intensive and unsuitable for pre-training with vast data. Considering the practical feasibility, we confine our exploration to the realm of memory replay methods.

\subsection{Key Factors in Memory Replay}
We have considered several potential design dimensions within the replay process:
\begin{itemize}[leftmargin=0pt]
\item \textbf{Replay Frequency.} Following~\citet{de2019episodic}, we match the size of our retrieved memory batches to our training batches. We execute a retrieval and gradient update every 100 steps, achieving an efficient 1\% replay rate.
\item \textbf{What to Store into Memory.} We consider strategies for memory sample storage: (1) including all samples encountered during pre-training, (2) prioritizing samples with entities, and (3) choosing high-loss samples that may be more susceptible to forgetting. Advanced selection methods are reserved for future research.
\item \textbf{Retrieve Strategy.} We've introduced two basic but impactful retrieval methods: random sampling and similarity-based sampling. Unlike~\citet{de2019episodic}, who used a pre-trained BERT~\citep{devlin2018bert} model for the similarity-based sampling, we opted for BM25~\citep{robertson2009probabilistic} for its efficiency~\citep{yao2022nlp}.
\item \textbf{Exit Mechanism.} Given the fixed intervals of memory replay, the number of replayable samples is inherently limited. Simple replay methods may lead to an imbalance in the samples being replayed, such as coincidentally focusing on a few samples every replay batch. Thus, limiting the maximum replay threshold of a sample may help.
\end{itemize}

\subsection{Experimental Settings}\label{table setting}

In the previous section, we used two datasets, A and B, to study the forgetting effect. Now, to mimic a realistic pre-training setup, we've mixed and shuffled A with B into one complete set. We trained GPT2 from scratch using this combined set. To measure forgetting across the dataset, we took 1/5 of A+B, selected samples with entities, and made an test set($\sim$ 200,000 samples). We then use the aforementioned 4 metrics to assess the results.

Although the ability to relearn past samples is beneficial, a drawback of the replay method is its increased training cost. Considering realistic feasibility and the need for simplicity, we select the following straightforward strategies, while leaving more sophisticated replay methods for future work:
\begin{itemize}[leftmargin=0pt]
    \item \textbf{Vanilla pre-training} The standard pre-training.
    \item \textbf{Upper Bound} We train from the above pre-training checkpoint on the test set, evaluating immediately to determine the model's peak memory retention.
    \item \textbf{BM25}. We leverage Elasticsearch~\citep{elasticsearch2018elasticsearch} to maintain a memory of all encountered samples. At designated replay intervals, we use the current batch as queries to search for previously seen similar data for replay.
    \item \textbf{BM25 + Samples with entities only}. During learning, we only keep samples with the presence of entities in our memory.
    \item \textbf{Focused Stochasticity: Constrained Entity Sampling with Exit Limit}. We shift from similarity-based retrieval to random sampling. We use the previously mentioned exit mechanism and exclude samples from the memory after they have been replayed 5 times.
    \item \textbf{Intensive Focused Stochasticity}: This variant of Focused Stochasticity intensifies the replay process, subjecting replayed samples to multiple epochs of learning. The idea behind this method and further details are elaborated in Section \ref{Periodic Intensive Replay}.
    
    Let $T_0$ denote the computational cost of vanilla pre-training, $T$ represent the interval between replays, and $f$ be the number of epochs conducted on the replay batch. The computational cost for the Intensive Focused Stochasticity method is $T_{\text{replay}} = (1+f/T)T_0$. We use $f = 5$ and $T=100$ in this experiment. Thus $T_{\text{replay}} = 1.05T_0$, which is affordable for practical use. More discussions are presented in Appendix~\ref{sec:discussions}.

\end{itemize}
\begin{table} [hbtp]
\setlength{\tabcolsep}{1.6mm}
\centering
\resizebox{\linewidth}{!}{%
    \begin{tabular}{l|cccc}
    \toprule[1pt]
    Method &PPL$_{\text{ent}}$ &M(f)$_{\text{ent}}$ & $M_{\text{ex}}$ ($\times 10^{-3}$)& $M_{\text{in}}$ ($\times 10^{-2}$)\\
    \midrule[1pt]
    Vanilla pre-training&26.03&0.4093&5.273&3.988\\
    Upper Bound&23.74&0.4182&14.46&4.162\\
    \hline
    BM25&27.95&0.4015&4.586&3.895\\
    BM25 + Samples with entities only&28.09&0.4013&4.575&3.941\\
    Focused Stochasticity&25.79&0.4101&\textbf{5.496}&3.980\\ 
    Intensive Focused Stochasticity&\textbf{25.40}&\textbf{0.4121}&5.450&\textbf{4.003}\\
    \bottomrule[1pt]
    \end{tabular}
}
\caption{Evaluation results for replay strategies.}
\label{tab:maintable}
\end{table}

\subsection{Effectiveness of Memory Replay}\label{table res}

We display the evaluation in Table \ref{tab:maintable}. The results indicates that similarity-based replay methods do not outperform the baseline, no matter if all samples or only those related to entities are kept in memory. This might be due to the strategies don't spread replay evenly; replaying all samples might focus too much on non-entity ones, while focusing only on entity-related samples could lead to too much attention on a specific subset, exaggerating the forgetting of other samples.

On the other hand, a simple sampling method improves upon the baseline, hinting that replay helps reduce forgetting during pre-training. Nevertheless, there's still a gap between the replay methods and the upperbound.

To further demonstrate the effectiveness of memory replay, we conducted an in-depth analysis of the impact of sample-level forgetting on the model's performance across common benchmark datasets. We utilized the following common benchmark datasets for our analysis: Hellaswag~\citep{zellers2019hellaswag}, MMLU~\citep{hendrycks2020measuring} and Winograd~\citep{levesque2012winograd}. We compared the zero-shot accuracy between the vanilla pre-training and our Intensive Focused Stochasticity.

\begin{table} [hbtp]
\setlength{\tabcolsep}{1.6mm}
\centering
\resizebox{\linewidth}{!}{%
    \begin{tabular}{l|cccc}
    \toprule[1pt]
    Method &Hellaswag&MMLU&Winograd&Avg.\\
    \midrule[1pt]
    Vanilla pre-training&27.46&\textbf{23.20}&53.47&34.71\\
    Intensive Focused Stochasticity&\textbf{27.75}&23.00&\textbf{55.68}&\textbf{35.48}\\
    \bottomrule[1pt]
    \end{tabular}
}
\caption{Results across common benchmark datasets.}
\label{tab:common_datasets}
\end{table}
The performance shows that Intensive Focused Stochasticity method is generally superior to the non-replay method. The MMLU dataset is relatively more difficult, and the results may be subject to disturbances. The results indicates that intensified memory replay methods offer improvements compared to the standard pre-training approach. Considering forgetting do help performance on downstream tasks.

\begin{tcolorbox}[leftrule=1.0mm,top=0.mm,bottom=0.0mm]
\textbf{Takeaway 3:} Our memory replay methods show potential in alleviating forgetting in the pre-training phase, while a gap persists relative to the upper bound, signifying the necessity for further research.
\end{tcolorbox}
\section{Explorations on Forgetting Curves}
In the preceding section, we demonstrated the efficacy of memory replay methods. Recognizing that traditional memory replay methods~\citep{de2019episodic, wang2020efficient} involve samples being learned uniformly and at equal intervals with low intensity. We now seek to explore the impact of replay learning on subsequent learning processes, as well as investigate factors such as the intensity of replay and the effects of periodic replay on learning curves. This exploration is motivated by the renowned forgetting curve from human psychology~\citep{loftus1985evaluating}, which underscores the link between the intensity of learning and the pace of forgetting.


We first focus on factors that we expect to manifest their influence on the model's forgetting curve. After an in-depth observation, we aim to apply the phenomena observed on the forgetting curve to guide the design of memory replay methods during pre-training. This approach is intended to refine and understand our strategies for combating forgetting, ensuring that they are informed by empirical insights into the model's learning dynamics.



\subsection{Setup}
We focus on two critical factors: (1) \textbf{Learning intensity's impact}: We explore the hypothesis that increased initial learning intensity may result in more robust memory retention, potentially flattening the forgetting curve. (2) \textbf{Memorability and memory durability}: We determine if challenging samples, post-intensive learning, remain at risk of forgetting during pre-training.
\begin{figure*}[t!]
  \centering
  \subfigure[PPL, low difficulty.]
  {
    \includegraphics[width=0.23\linewidth]{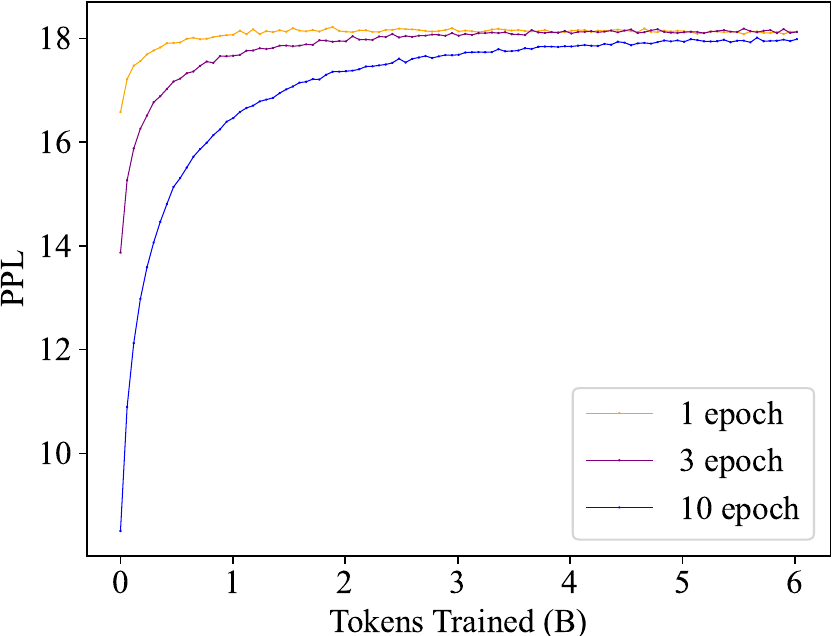}
  }
  \subfigure[M(f), low difficulty.]
  {
    \includegraphics[width=0.23\linewidth]{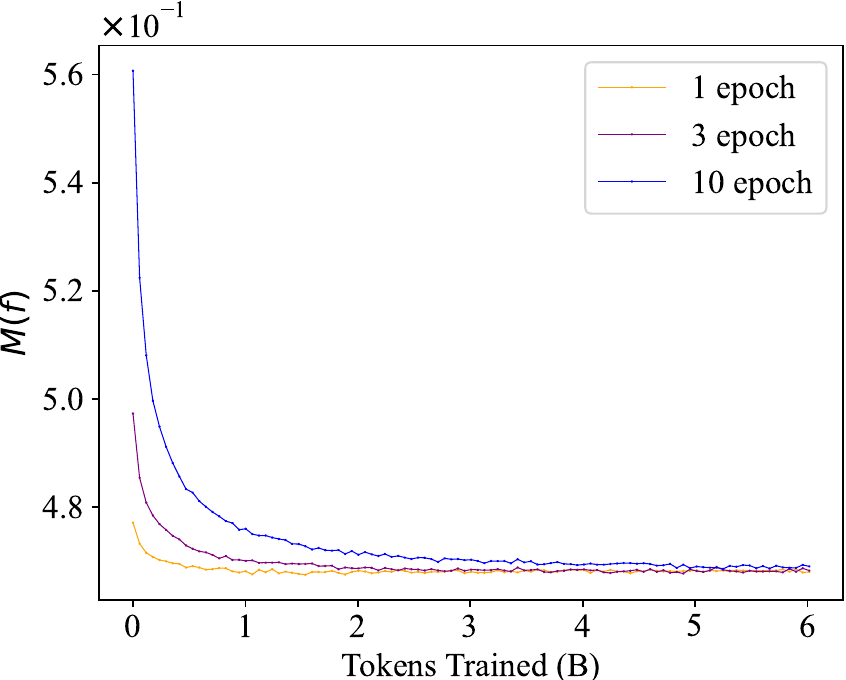}
  }
  \subfigure[$M_{\text{in}}$, low difficulty.]
  {
    \includegraphics[width=0.23\linewidth]{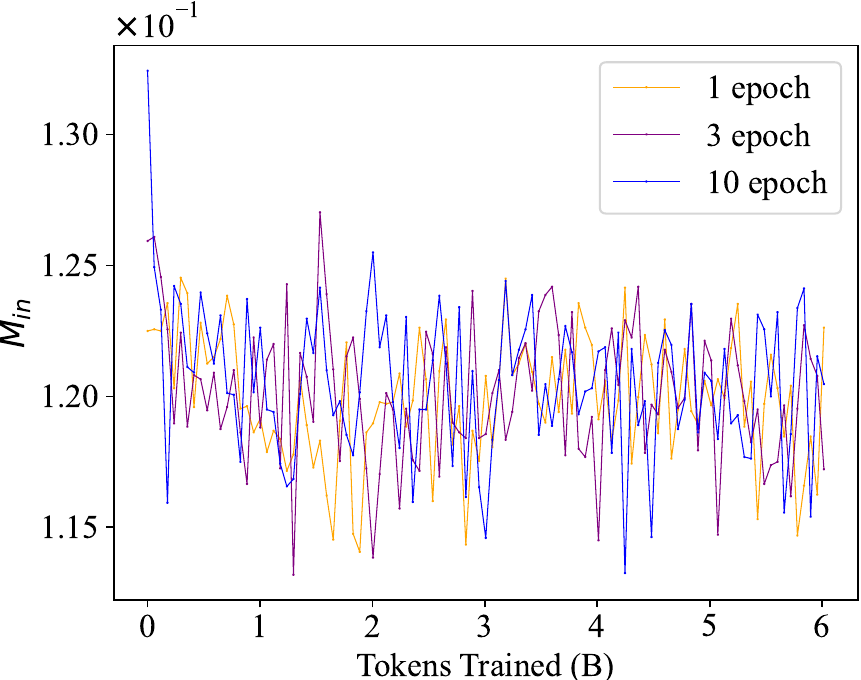}
  }
  \subfigure[$M_{\text{ex}}$, low difficulty.]
  {
    \includegraphics[width=0.23\linewidth]{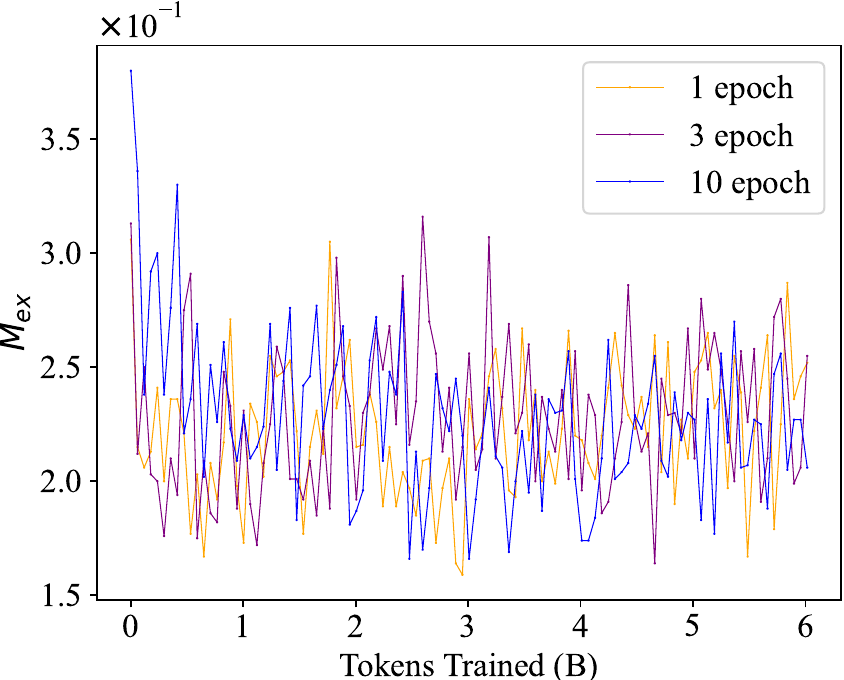}
  }
  
  \subfigure[PPL, high difficulty.]
  {
    \includegraphics[width=0.23\linewidth]{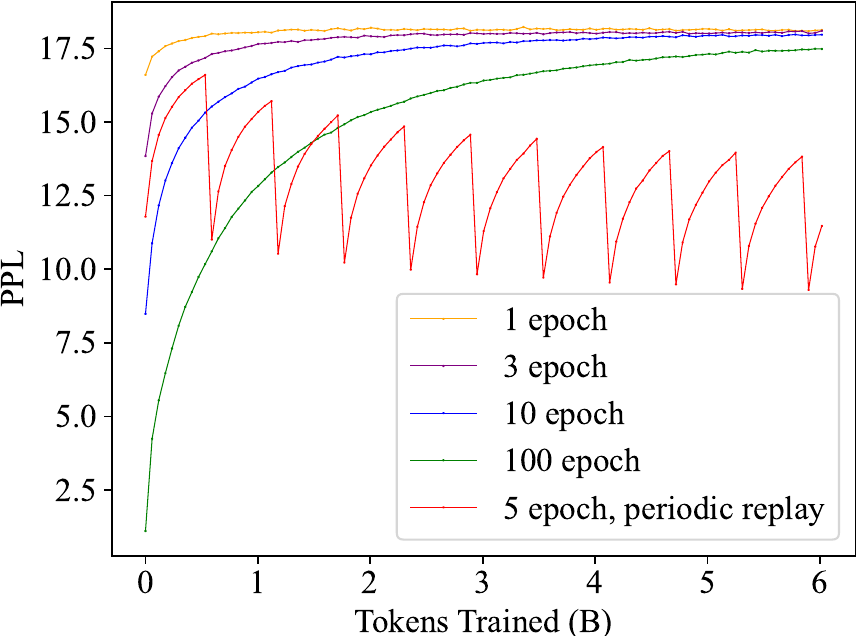}
  }  
  \subfigure[M(f), high difficulty.]
  {
    \includegraphics[width=0.23\linewidth]{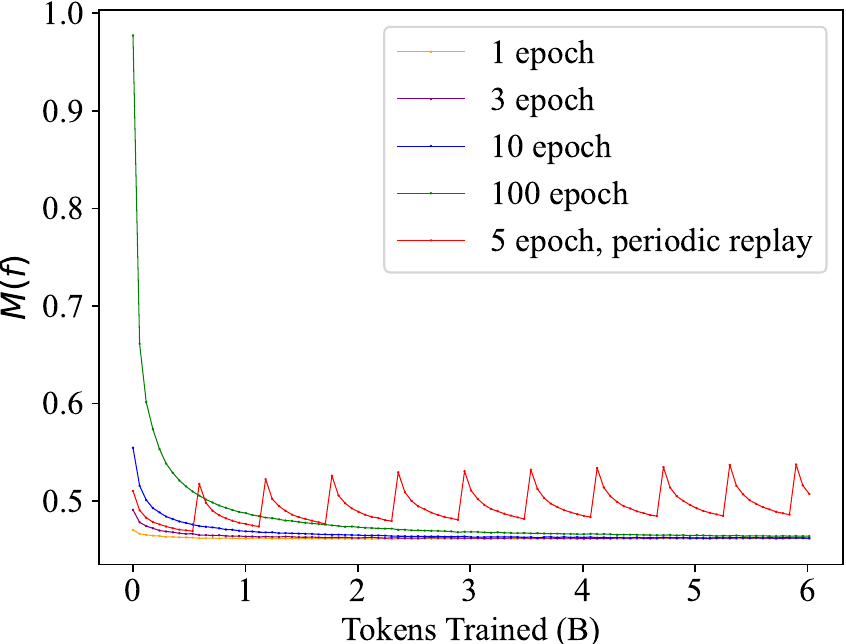}
  }
  \subfigure[$M_{\text{in}}$, high difficulty.]
  {
    \includegraphics[width=0.23\linewidth]{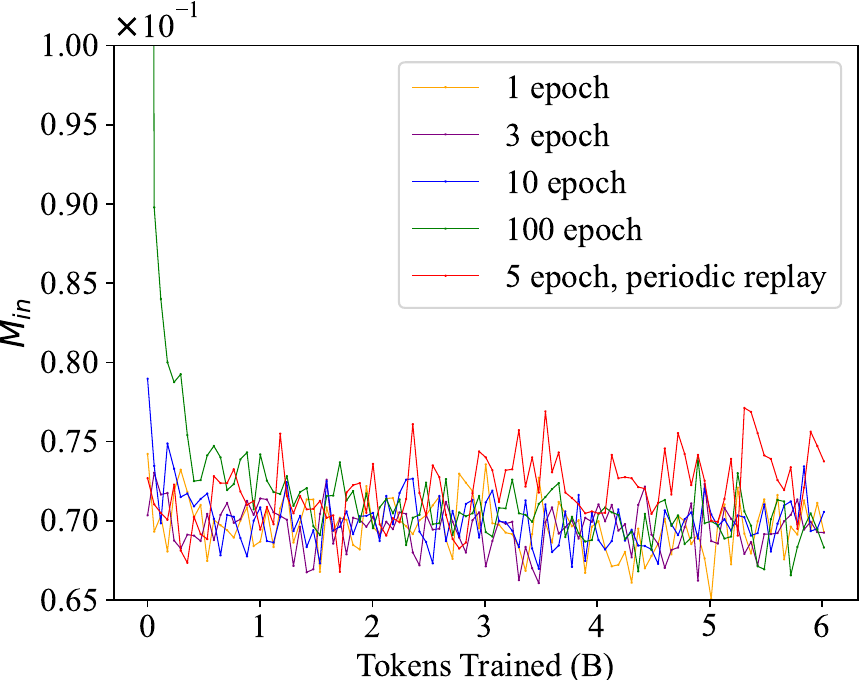}
  }
  \subfigure[$M_{\text{ex}}$, high difficulty.]
  {
    \includegraphics[width=0.23\linewidth]{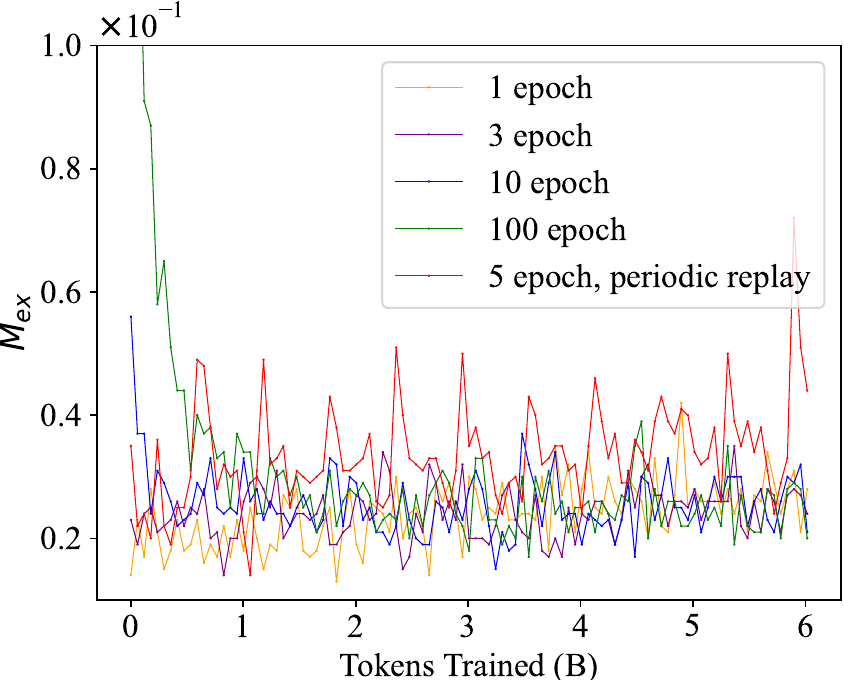}
  }
  
  \caption{Forgetting curves on samples categorized by difficulty level. After sufficiently training, experiments with varying degrees of replay intensity tend to converge, while there remains a gap between methods with higher and lower replay intensities. \textbf{Our key experiment}, periodic replay method (\textbf{red}) demonstrates the ability to achieve continuous performance improvement across the entire learning curve with a smaller computational cost. Remarkably, even at the end of the curve, the upper and lower bounds of the periodic replay method remain consistently better.
}
  \label{fig:forget curve}
\end{figure*}

To tackle these inquiries, we first select samples related to entities of interest. After the model undergoes an initial epoch of pre-training, we subject these samples to intensive training across several epochs. The purpose of the initial pre-training epoch is to ensure the model reaches a basic level of language proficiency. This step is crucial to prevent general language ability improvements from confounding the experiment, allowing for a clear focus on the forgetting phenomenon rather than overall enhancement.


Post the intensive learning phase, these entity-related samples serve as our test set. As we proceed with pre-training, we continuously assess this set using our established metrics to monitor the forgetting curve. This ongoing evaluation allows us to track how the memory of these samples evolves and to understand the interplay between initial learning intensity and long-term retention within the context of pre-training. For further details on the experimental design, please refer to the Appendix~\ref{appendix forget curve}.

\subsection{Results on LLMs' Forgetting Curves}
\label{forget curve}
\subsubsection{Forgetting Curves}
As shown in Figure \ref{fig:forget curve}, experiments indicate that (1) a significant decline is still observed even when the dataset used for subsequent training is \textit{identical and uniformly distributed} to the source of the data in the initial epoch of pre-training. This reinforces our conclusions presented in Section \ref{new metric}, reflecting that \textbf{even under an identical data distribution}, forgetting is still remarkably pronounced. (2) higher initial learning intensity results in better performance across various metrics, yet as further pre-training occurs, the results from experiments with lower initial learning intensity tend to catch up. This pattern mirrors human learning curves~\citep{loftus1985evaluating}, and we offer a detailed comparison in Appendix \ref{appendix human forget curve}. (3) Over the learning period, a divergence is observed; experiments with a very high initial learning intensity \textbf{maintain a gap} compared to those with a lower initial intensity. This gap is more pronounced for less difficult data. This suggests that data that are more difficult to memorize benefit from more intensive learning to achieve enhanced memory retention.

\subsubsection{Periodic Intensive Replay}\label{Periodic Intensive Replay}
Building on findings above, we recall the intuition that human can reduce forgetting through periodic, intense learning. We aim to (1) assess its impact on a model's forgetting curve, and (2) determine if this can enhance previous memory replay methods. To delve deeper into these effects, we focused our experiments on the more challenging samples. After the initial phase of high-intensity learning, we introduced a replay process in the ongoing pre-training. This process involves revisiting the samples every 1000 steps, with each replay session consisting of 5 epochs of learning.

In this experiment, the replay intervals were relatively large, which was acceptable in terms of efficiency. Moreover, the \textbf{periodic replay method outperformed the baseline}. Although there was a temporary decline after each replay, the overall performance improves over time. We discovered that periodic, high-intensity replay on the forgetting curve leads to an enhancement of both the upper and lower bounds. Moreover, this approach proved more effective and cost-efficient than directly replay with 100 epochs.

\subsubsection{Intensive Replay for Pre-training}
Thus, we believe that such human-like strategies could guide the design of replay mechanisms. To test this hypothesis, we conducted an experiment and enhanced the Focused Stochasticity method in Section \ref{table setting}. Specifically, we intensified the learning process for each replay batch, with each batch undergoing 5 epochs of learning. The approach, referred to as Intensive Focused Stochasticity, has been included in Table~\ref{tab:maintable} for ease of comparison with other methods. Additionally, its performance on general downstream tasks is presented in Table~\ref{tab:common_datasets}. The results indicate that Intensive Focused Stochasticity outperforms vanilla pre-training across all metrics, suggesting its efficacy in mitigating forgetting during pre-training.
\begin{tcolorbox}[leftrule=1.0mm,top=0.mm,bottom=0.0mm]
\textbf{Takeaway 4:} The forgetting patterns of LLMs suggest that periodic, intensive replay could be key to mitigating memory loss. Experiments of intensified memory replay conducted during the pre-training phase also confirmed this point.
\end{tcolorbox}
\section{Conclusion and Future Work}
We aspire to capture the industry's attention and stimulate optimization efforts regarding the often-overlooked potential danger within model development. Our research sheds new light on catastrophic forgetting in LLMs during pre-training. We scrutinized traditional metrics, introduced novel ones for a clearer analysis of forgetting, and proposed memory-replay techniques to bolster knowledge retention. Additionally, we explored the forgetting curve post-intense, short-term learning, uncovering similarities with human memory decay, offering insights into information retention dynamics.


\section{Limitations}

Our research into the occurrence of catastrophic forgetting during the pre-training of Large Language Models, though innovative, faces significant computational constraints. The necessity for a substantial computational resource, such as 10,000 GPU hours on 8 NVIDIA A100 GPUs equipped with 40 GiB of VRAM, presents a considerable barrier. The core contribution of our work is to emphasize and delve deeply into an often-overlooked potential danger, with the hope of drawing the industry's attention to and optimizing the issue of pre-training forgetting in models.

Informed by the scaling law~\citep{kaplan2020scaling}, we recognize that our findings from a smaller model may provide valuable insights for larger-scale experiments. This framework indicates that our study could contribute to the design of future research, acknowledging the limitations in scaling our results.

Our approach to memory replay has shown potential in alleviating catastrophic forgetting, but there is still room for improvement in terms of its effectiveness. Our investigation did not delve deeply into the granular effects of each variable on the experimental outcomes. The complexity of memory replay mechanisms requires a more nuanced analysis to fully understand how different factors interplay and influence the results.

Additionally, the concentrated learning of memory replay, while beneficial, may engender trade-offs that affect the model's generalizability. We hypothesize that the focused emphasis of certain data subsets could lead to a diminished capacity for the model to adapt to tasks beyond the focused areas, such as numerical data processing or other cognitively distinct downstream tasks.

We recognize that forgetting in pre-training differs from that in SFT, each requiring distinct metrics and methods for mitigation. Yet, there are connections between them. In future work, we also aim to explore the impact of our methods on forgetting in downstream tasks.

Despite these limitations, our study exemplifies the scientific endeavor to confront complex problems with rigor and without reservation. Our work is a courageous step towards understanding the intricate processes of memory retention and forgetting in LLMs, reflecting a sincere commitment to advancing our collective knowledge, even in the face of substantial challenges.

\bibliography{custom}

\begin{thebibliography}{40}
\expandafter\ifx\csname natexlab\endcsname\relax\def\natexlab#1{#1}\fi

\bibitem[{Allingham et~al.(2021)Allingham, Wenzel, Mariet, Mustafa, Puigcerver, Houlsby, Jerfel, Fortuin, Lakshminarayanan, Snoek et~al.}]{allingham2021sparse}
James~Urquhart Allingham, Florian Wenzel, Zelda~E Mariet, Basil Mustafa, Joan Puigcerver, Neil Houlsby, Ghassen Jerfel, Vincent Fortuin, Balaji Lakshminarayanan, Jasper Snoek, et~al. 2021.
\newblock Sparse moes meet efficient ensembles.
\newblock \emph{arXiv preprint arXiv:2110.03360}.

\bibitem[{Biderman et~al.(2023{\natexlab{a}})Biderman, Prashanth, Sutawika, Schoelkopf, Anthony, Purohit, and Raf}]{biderman2023emergent}
Stella Biderman, USVSN~Sai Prashanth, Lintang Sutawika, Hailey Schoelkopf, Quentin Anthony, Shivanshu Purohit, and Edward Raf. 2023{\natexlab{a}}.
\newblock Emergent and predictable memorization in large language models.
\newblock \emph{arXiv preprint arXiv:2304.11158}.

\bibitem[{Biderman et~al.(2023{\natexlab{b}})Biderman, Schoelkopf, Anthony, Bradley, O’Brien, Hallahan, Khan, Purohit, Prashanth, Raff et~al.}]{biderman2023pythia}
Stella Biderman, Hailey Schoelkopf, Quentin~Gregory Anthony, Herbie Bradley, Kyle O’Brien, Eric Hallahan, Mohammad~Aflah Khan, Shivanshu Purohit, USVSN~Sai Prashanth, Edward Raff, et~al. 2023{\natexlab{b}}.
\newblock Pythia: A suite for analyzing large language models across training and scaling.
\newblock In \emph{International Conference on Machine Learning}. PMLR.

\bibitem[{Bourtoule et~al.(2021)Bourtoule, Chandrasekaran, Choquette-Choo, Jia, Travers, Zhang, Lie, and Papernot}]{bourtoule2021machine}
Lucas Bourtoule, Varun Chandrasekaran, Christopher~A Choquette-Choo, Hengrui Jia, Adelin Travers, Baiwu Zhang, David Lie, and Nicolas Papernot. 2021.
\newblock Machine unlearning.
\newblock In \emph{2021 IEEE Symposium on Security and Privacy (SP)}. IEEE.

\bibitem[{Chang et~al.(2024)Chang, Park, Ye, Yang, Seo, Chang, and Seo}]{chang2024large}
Hoyeon Chang, Jinho Park, Seonghyeon Ye, Sohee Yang, Youngkyung Seo, Du-Seong Chang, and Minjoon Seo. 2024.
\newblock How do large language models acquire factual knowledge during pretraining?
\newblock \emph{arXiv preprint arXiv:2406.11813}.

\bibitem[{Chen et~al.(2022)Chen, Zhang, Wang, Backes, Humbert, and Zhang}]{chen2022graph}
Min Chen, Zhikun Zhang, Tianhao Wang, Michael Backes, Mathias Humbert, and Yang Zhang. 2022.
\newblock Graph unlearning.
\newblock In \emph{Proceedings of the 2022 ACM SIGSAC conference on computer and communications security}.

\bibitem[{Craig et~al.(1972)Craig, Sternthal, and Olshan}]{craig1972effect}
C~Samuel Craig, Brian Sternthal, and Karen Olshan. 1972.
\newblock The effect of overlearning on retention.
\newblock \emph{Journal of General Psychology}.

\bibitem[{de~Masson~D'Autume et~al.(2019)de~Masson~D'Autume, Ruder, Kong, and Yogatama}]{de2019episodic}
Cyprien de~Masson~D'Autume, Sebastian Ruder, Lingpeng Kong, and Dani Yogatama. 2019.
\newblock Episodic memory in lifelong language learning.
\newblock \emph{NeurIPS}.

\bibitem[{Devlin et~al.(2018)Devlin, Chang, Lee, and Toutanova}]{devlin2018bert}
Jacob Devlin, Ming-Wei Chang, Kenton Lee, and Kristina Toutanova. 2018.
\newblock Bert: Pre-training of deep bidirectional transformers for language understanding.
\newblock \emph{arXiv preprint arXiv:1810.04805}.

\bibitem[{Elasticsearch(2018)}]{elasticsearch2018elasticsearch}
BV~Elasticsearch. 2018.
\newblock Elasticsearch.
\newblock \emph{software], version}.

\bibitem[{Foundation()}]{wikidump}
Wikimedia Foundation.
\newblock \href {https://dumps.wikimedia.org} {Wikimedia downloads}.

\bibitem[{Gao et~al.(2020)Gao, Biderman, Black, Golding, Hoppe, Foster, Phang, He, Thite, Nabeshima et~al.}]{gao2020pile}
Leo Gao, Stella Biderman, Sid Black, Laurence Golding, Travis Hoppe, Charles Foster, Jason Phang, Horace He, Anish Thite, Noa Nabeshima, et~al. 2020.
\newblock The {P}ile: An 800{GB} dataset of diverse text for language modeling.
\newblock \emph{arXiv preprint arXiv:2101.00027}.

\bibitem[{Gokaslan et~al.(2019)Gokaslan, Cohen, Pavlick, and Tellex}]{Gokaslan2019OpenWeb}
Aaron Gokaslan, Vanya Cohen, Ellie Pavlick, and Stefanie Tellex. 2019.
\newblock Openwebtext corpus.
\newblock \url{http://Skylion007.github.io/OpenWebTextCorpus}.

\bibitem[{Gunasekar et~al.(2023)Gunasekar, Zhang, Aneja, Mendes, Del~Giorno, Gopi, Javaheripi, Kauffmann, de~Rosa, Saarikivi et~al.}]{gunasekar2023textbooks}
Suriya Gunasekar, Yi~Zhang, Jyoti Aneja, Caio C{\'e}sar~Teodoro Mendes, Allie Del~Giorno, Sivakanth Gopi, Mojan Javaheripi, Piero Kauffmann, Gustavo de~Rosa, Olli Saarikivi, et~al. 2023.
\newblock Textbooks are all you need.
\newblock \emph{arXiv preprint arXiv:2306.11644}.

\bibitem[{Gupta et~al.(2023)Gupta, Th{\'e}rien, Ibrahim, Richter, Anthony, Belilovsky, Rish, and Lesort}]{gupta2023continual}
Kshitij Gupta, Benjamin Th{\'e}rien, Adam Ibrahim, Mats~L Richter, Quentin Anthony, Eugene Belilovsky, Irina Rish, and Timoth{\'e}e Lesort. 2023.
\newblock Continual pre-training of large language models: How to (re) warm your model?
\newblock \emph{arXiv preprint arXiv:2308.04014}.

\bibitem[{Hendrycks et~al.(2020)Hendrycks, Burns, Basart, Zou, Mazeika, Song, and Steinhardt}]{hendrycks2020measuring}
Dan Hendrycks, Collin Burns, Steven Basart, Andy Zou, Mantas Mazeika, Dawn Song, and Jacob Steinhardt. 2020.
\newblock Measuring massive multitask language understanding.
\newblock \emph{arXiv preprint arXiv:2009.03300}.

\bibitem[{Jin et~al.(2021)Jin, Zhang, Zhu, Xiao, Li, Wei, Arnold, and Ren}]{jin2021lifelong}
Xisen Jin, Dejiao Zhang, Henghui Zhu, Wei Xiao, Shang-Wen Li, Xiaokai Wei, Andrew Arnold, and Xiang Ren. 2021.
\newblock Lifelong pretraining: Continually adapting language models to emerging corpora.
\newblock \emph{arXiv preprint arXiv:2110.08534}.

\bibitem[{Kaplan et~al.(2020)Kaplan, McCandlish, Henighan, Brown, Chess, Child, Gray, Radford, Wu, and Amodei}]{kaplan2020scaling}
Jared Kaplan, Sam McCandlish, Tom Henighan, Tom~B Brown, Benjamin Chess, Rewon Child, Scott Gray, Alec Radford, Jeffrey Wu, and Dario Amodei. 2020.
\newblock Scaling laws for neural language models.
\newblock \emph{arXiv preprint arXiv:2001.08361}.

\bibitem[{Levesque et~al.(2012)Levesque, Davis, and Morgenstern}]{levesque2012winograd}
Hector Levesque, Ernest Davis, and Leora Morgenstern. 2012.
\newblock The winograd schema challenge.
\newblock In \emph{Thirteenth international conference on the principles of knowledge representation and reasoning}.

\bibitem[{Loftus(1985)}]{loftus1985evaluating}
Geoffrey~R Loftus. 1985.
\newblock Evaluating forgetting curves.
\newblock \emph{Journal of Experimental Psychology: Learning, Memory, and Cognition}.

\bibitem[{Luo et~al.(2023)Luo, Yang, Meng, Li, Zhou, and Zhang}]{luo2023empirical}
Yun Luo, Zhen Yang, Fandong Meng, Yafu Li, Jie Zhou, and Yue Zhang. 2023.
\newblock An empirical study of catastrophic forgetting in large language models during continual fine-tuning.
\newblock \emph{arXiv preprint arXiv:2308.08747}.

\bibitem[{McCloskey and Cohen(1989)}]{mccloskey1989catastrophic}
Michael McCloskey and Neal~J Cohen. 1989.
\newblock Catastrophic interference in connectionist networks: The sequential learning problem.
\newblock In \emph{Psychology of learning and motivation}. Elsevier.

\bibitem[{Penedo et~al.(2023)Penedo, Malartic, Hesslow, Cojocaru, Cappelli, Alobeidli, Pannier, Almazrouei, and Launay}]{penedo2023refinedweb}
Guilherme Penedo, Quentin Malartic, Daniel Hesslow, Ruxandra Cojocaru, Alessandro Cappelli, Hamza Alobeidli, Baptiste Pannier, Ebtesam Almazrouei, and Julien Launay. 2023.
\newblock The refinedweb dataset for falcon llm: outperforming curated corpora with web data, and web data only.
\newblock \emph{arXiv preprint arXiv:2306.01116}.

\bibitem[{Qin et~al.(2022)Qin, Zhang, Lin, Liu, Li, Sun, and Zhou}]{qin2022elle}
Yujia Qin, Jiajie Zhang, Yankai Lin, Zhiyuan Liu, Peng Li, Maosong Sun, and Jie Zhou. 2022.
\newblock Elle: Efficient lifelong pre-training for emerging data.
\newblock \emph{arXiv preprint arXiv:2203.06311}.

\bibitem[{Radford et~al.(2019)Radford, Wu, Child, Luan, Amodei, Sutskever et~al.}]{radford2019language}
Alec Radford, Jeffrey Wu, Rewon Child, David Luan, Dario Amodei, Ilya Sutskever, et~al. 2019.
\newblock Language models are unsupervised multitask learners.
\newblock \emph{OpenAI blog}.

\bibitem[{Rajbhandari et~al.(2020)Rajbhandari, Rasley, Ruwase, and He}]{rajbhandari2020zero}
Samyam Rajbhandari, Jeff Rasley, Olatunji Ruwase, and Yuxiong He. 2020.
\newblock Zero: Memory optimizations toward training trillion parameter models.
\newblock In \emph{SC20: International Conference for High Performance Computing, Networking, Storage and Analysis}. IEEE.

\bibitem[{Ratcliff(1990)}]{ratcliff1990connectionist}
Roger Ratcliff. 1990.
\newblock Connectionist models of recognition memory: constraints imposed by learning and forgetting functions.
\newblock \emph{Psychological review}.

\bibitem[{Robertson et~al.(2009)Robertson, Zaragoza et~al.}]{robertson2009probabilistic}
Stephen Robertson, Hugo Zaragoza, et~al. 2009.
\newblock The probabilistic relevance framework: Bm25 and beyond.
\newblock \emph{Foundations and Trends{\textregistered} in Information Retrieval}.

\bibitem[{Soboleva et~al.(2023)Soboleva, Al-Khateeb, Myers, Steeves, Hestness, and Dey}]{cerebras2023slimpajama}
Daria Soboleva, Faisal Al-Khateeb, Robert Myers, Jacob~R Steeves, Joel Hestness, and Nolan Dey. 2023.
\newblock \href {https://huggingface.co/datasets/cerebras/SlimPajama-627B} {{SlimPajama: A 627B token cleaned and deduplicated version of RedPajama}}.

\bibitem[{Sorscher et~al.(2022)Sorscher, Geirhos, Shekhar, Ganguli, and Morcos}]{sorscher2022beyond}
Ben Sorscher, Robert Geirhos, Shashank Shekhar, Surya Ganguli, and Ari Morcos. 2022.
\newblock Beyond neural scaling laws: beating power law scaling via data pruning.
\newblock \emph{NeurIPS}.

\bibitem[{Sun et~al.(2020)Sun, Wang, Li, Feng, Tian, Wu, and Wang}]{sun2020ernie}
Yu~Sun, Shuohuan Wang, Yukun Li, Shikun Feng, Hao Tian, Hua Wu, and Haifeng Wang. 2020.
\newblock Ernie 2.0: A continual pre-training framework for language understanding.
\newblock In \emph{AAAI}.

\bibitem[{Tirumala et~al.(2022)Tirumala, Markosyan, Zettlemoyer, and Aghajanyan}]{tirumala2022memorization}
Kushal Tirumala, Aram Markosyan, Luke Zettlemoyer, and Armen Aghajanyan. 2022.
\newblock Memorization without overfitting: Analyzing the training dynamics of large language models.
\newblock \emph{NeurIPS}.

\bibitem[{Toneva et~al.(2018)Toneva, Sordoni, Combes, Trischler, Bengio, and Gordon}]{toneva2018empirical}
Mariya Toneva, Alessandro Sordoni, Remi Tachet~des Combes, Adam Trischler, Yoshua Bengio, and Geoffrey~J Gordon. 2018.
\newblock An empirical study of example forgetting during deep neural network learning.
\newblock \emph{arXiv preprint arXiv:1812.05159}.

\bibitem[{Wang et~al.(2023{\natexlab{a}})Wang, Liu, Yue, Tang, Zhang, Jiayang, Yao, Gao, Hu, Qi et~al.}]{wang2023survey}
Cunxiang Wang, Xiaoze Liu, Yuanhao Yue, Xiangru Tang, Tianhang Zhang, Cheng Jiayang, Yunzhi Yao, Wenyang Gao, Xuming Hu, Zehan Qi, et~al. 2023{\natexlab{a}}.
\newblock Survey on factuality in large language models: Knowledge, retrieval and domain-specificity.
\newblock \emph{arXiv preprint arXiv:2310.07521}.

\bibitem[{Wang et~al.(2023{\natexlab{b}})Wang, Chen, Ge, Xia, Bao, Zheng, Zhang, Gui, and Huang}]{wang2023orthogonal}
Xiao Wang, Tianze Chen, Qiming Ge, Han Xia, Rong Bao, Rui Zheng, Qi~Zhang, Tao Gui, and Xuanjing Huang. 2023{\natexlab{b}}.
\newblock Orthogonal subspace learning for language model continual learning.
\newblock \emph{arXiv preprint arXiv:2310.14152}.

\bibitem[{Wang et~al.(2020)Wang, Mehta, P{\'o}czos, and Carbonell}]{wang2020efficient}
Zirui Wang, Sanket~Vaibhav Mehta, Barnab{\'a}s P{\'o}czos, and Jaime Carbonell. 2020.
\newblock Efficient meta lifelong-learning with limited memory.
\newblock \emph{arXiv preprint arXiv:2010.02500}.

\bibitem[{Wu et~al.(2024)Wu, Gan, Ge, Lu, Wang, Feng, Luo, and Shan}]{wu2024llama}
Chengyue Wu, Yukang Gan, Yixiao Ge, Zeyu Lu, Jiahao Wang, Ye~Feng, Ping Luo, and Ying Shan. 2024.
\newblock Llama pro: Progressive llama with block expansion.
\newblock \emph{arXiv preprint arXiv:2401.02415}.

\bibitem[{Wu et~al.(2020)Wu, Dobriban, and Davidson}]{wu2020deltagrad}
Yinjun Wu, Edgar Dobriban, and Susan Davidson. 2020.
\newblock Deltagrad: Rapid retraining of machine learning models.
\newblock In \emph{International Conference on Machine Learning}. PMLR.

\bibitem[{Yao et~al.(2022)Yao, Zheng, Yang, and Yang}]{yao2022nlp}
Xingcheng Yao, Yanan Zheng, Xiaocong Yang, and Zhilin Yang. 2022.
\newblock Nlp from scratch without large-scale pretraining: A simple and efficient framework.
\newblock PMLR.

\bibitem[{Zellers et~al.(2019)Zellers, Holtzman, Bisk, Farhadi, and Choi}]{zellers2019hellaswag}
Rowan Zellers, Ari Holtzman, Yonatan Bisk, Ali Farhadi, and Yejin Choi. 2019.
\newblock Hellaswag: Can a machine really finish your sentence?
\newblock \emph{arXiv preprint arXiv:1905.07830}.

\end{thebibliography}
\newpage
\clearpage
\appendix
\section{TL;DR: Main Contributions}
In this work, our focus is on exploring an issue that developers and researchers in the industry have frequently noticed: large models, despite their widespread use, are susceptible to errors in factual domains, particularly regarding entity-related information~\citep{wang2023survey}. While the erosion of knowledge retention during pre-training is acknowledged, no previous work has addressed the issue of forgetting in pre-training, nor provided a clear definition, analysis, or methods to address it. Our core contributions in this work are:
\begin{itemize}[leftmargin=0pt]
    \item We are the first to identify the problem of forgetting during pre-training.
    \item Within an affordable computational range, we conducted dozens of experiments to carefully explore the existence of the pre-training forgetting issue, the metrics for measurement, the forgetting curve, and the design of replay methods guided by the forgetting curve to provide feasible methods for mitigating pre-training forgetting.
\end{itemize}

Although the issue of forgetting is important and has been extensively studied during the SFT phase, no one is willing to tackle such a challenging problem in pre-training. The pretrain data is extremely vast and complex, inherently containing thousands of tasks. It cannot be characterized by task-level metrics, and such metrics also cannot reflect the general factual forgetting. Moreover, representing the forgetting of task-specific capabilities is too vague and elusive. In pre-training, most efforts have focused on synthetic data~\citep{gunasekar2023textbooks} and model structures~\citep{allingham2021sparse}, with too little research on the phenomenon itself.

We hope that the explorations and conclusions presented in this paper can facilitate the design of pre-training in the industry. We also aim to conduct experiments on larger models and more diverse datasets to provide more detailed conclusions.
\section{Further Discussions on Pre-training Forgetting}

In this section, we discuss the intuition and methodology behind the paper, as well as potential issues.
\begin{enumerate}
\item \textbf{Why should we be concerned about model forgetting at the sample level during pre-training?} 

Developers and researchers have frequently observed that large models, despite their extensive deployment, are prone to errors in factual domains, especially concerning entity-related information~\citep{wang2023survey}. These discrepancies can substantially affect user perception and trust. However, there is a scarcity of research on the influence of learning during the pre-training phase on this type of information, and even less on how models remember and forget information during pre-training. The phenomenon of sample-level forgetting in pre-training is also difficult to define clearly, analyze, and further explore.
\item \textbf{How should we understand entity-related metrics, and why is it important to focus on forgetting at the entity level?}

(1) Forgetting across the entire pre-training dataset is extremely difficult to define and study, hence we concentrate on a specific subset. Errors related to entity information are easily noticeable in model applications and significantly impact user experience. (2) Beyond the model's memory of entity information, we also consider its capabilities during pre-training, especially since the Supervised Fine-Tuning (SFT) phase places more emphasis on instructional data. This phase enhances the model's competencies for downstream tasks, and we see it as a stage for augmenting the model's capabilities. Therefore, we believe the pre-training phase should place greater emphasis on exploring entity information. (3) In Section \ref{2.2}, we demonstrate that overall data forgetting is hard to evaluate, as there is no clear decline in model performance when switching training data (we deliberately selected parts of data from A to ensure minimal repetition in B), and almost no change in metrics is observed during the switch. Instead, during training in B, the model's capabilities continue to improve, even surpassing the metrics achieved during training in A, which contradicts the intuition of forgetting. PPL does not intuitively reflect the model's forgetting; in contrast, the metrics concentrated on entities show significant changes on entity-related data, with almost no recovery, facilitating the direct study of the forgetting phenomenon.

\item \textbf{Why the proposed metrics better reflect forgetting? Might the decreased performance on the metric be attributed to the application of a more stringent metric?}

Attempting to identify the phenomenon of forgetting during pre-training and to indicate it with a reasonably sound metric poses a considerable challenge. However, this question has never been explored in the past. We have extensively reviewed previous work and have adopted the PPL and M(f) metrics, while also proposing two novel metrics.

The A and B datasets in Section~\ref{new metric}, as general pre-training datasets, show no significant differences in text style. Besides, in Section~\ref{forget curve}, we showed that a significant decline in metrics is still observed, even the dataset used for subsequent training is identical and uniformly distributed to the source of the data in the initial epoch of pre-training. This indicates that forgetting detected by our metrics does not stem from a shift in text styles.

Regarding the difficulty of metrics, in the experiment shown in Figure~\ref{fig:result}, we observe that even metrics that are simple by design, such as PPL and M(f), \textbf{show a significant decline}. This suggests that the forgetting phenomenon is unrelated to the difficulty of the metric. Besides, for M(f), which involves calculating the accuracy of the subsequent 32 tokens for each decoded token using teacher forcing, it is not simpler. However, we can see that PPL and M(f) slowly recover during subsequent training, indicating they are not sensitive enough to capture the forgetting phenomenon. While the $M_{ex}$ and $M_{in}$, though more complex, are more sensitive. We believe that by combining a range of metrics with varying degrees of design complexity and sensitivity, we can provide as comprehensive a portrayal of the phenomenon of forgetting as possible.

\item \textbf{Since the model may leak verbatim sequences of personal information, is sample-level forgetting harmful?}

In our study, we focus on learning and the retention of factual information related to entities, which models should not forget and that is prevalent in the pre-training data. We diverge from concerns about leaking verbatim personal information. There is extensive literature on machine unlearning~\citep{wu2020deltagrad,bourtoule2021machine,chen2022graph}, which typically addresses scenarios involving privacy protection and changes in user information. These scenarios fall outside the scope of our work, although our research might offer insights into the design of machine unlearning methods.
\item \textbf{Is this study primarily addressing hallucinations, or is it actually more focused on the model's tendency to forget entity-related information rather than producing false outputs?}

Our research concentrates on the model's inclination to forget information pertaining to entities, diverging from the generation of erroneous outputs, commonly known as hallucinations. However, it is true that our work offers a perspective on the concept of hallucinations, where the two newly designed metrics, \( M_{\text{ex}} \) and \( M_{\text{in}} \), can be interpreted as potential false negatives and false positives in the pre-training model's responses: the model, given relevant information, fails to identify the correct entity; or the model provides an entity and some information but is unable to supply the related context.
\item \textbf{Should we expect an LLM to reproduce exact training text, given it's not a lossless compression model?}

In our study, we do not anticipate LLMs to reproduce the exact training text. Specifically, our \( M_{\text{ex}} \) metric solely assesses whether the ground truth entity is included in the output; while capturing the formalization of information related to the entity presents challenges. For the \( M_{\text{in}} \) metric, we follow the design of~\citet{biderman2023emergent}, calculating accuracy for each token. We consider that alternative design schemes might be possible, such as utilizing a BERT model~\citep{devlin2018bert} to calculate the similarity between the generated tokens and the ground truth tokens. We have reserved this exploration for future research.
\item \textbf{Analysis of computational costs for replay methods.} To discuss the computational cost of replay methods, let $T_0$ denote the computational cost of vanilla pre-training, $T$ represent the interval between replays, and $f$ be the number of epochs conducted on the replay batch. $(1+f/T)T_0$. Every $T$ training steps, the model gets a replay batch and undergoes f epochs of training on that batch. Therefore, training \( T \) batches of unique data, replay methods necessitate \( T + f \) steps of training, whereas vanilla pre-training requires training with just \( T \) batches. This indicates that the computational cost for the Intensive Focused Stochasticity method is $T_{\text{replay}} = (1+f/T)T_0$. Setting $f = 1$, the Intensive Focused Stochasticity will degenerate to Focused Stochasticity. For instance, if \( f = 5 \), \( T = 100 \), and \( T_{\text{replay}} = 1.05 T_0 \), such computational cost is deemed acceptable.

\end{enumerate}

\section{Setup Details}
\label{sec:discussions}

In this section, we outline our experimental setup. We selected a batch size of 576, informed by our use of 8 NVIDIA A100 GPUs with 40 GiB VRAM, and aligned with GPT-2`s~\citep{radford2019language} hyperparameter recommendations for optimal performance on our hardware configuration. A consistent sequence length of 1024 was applied across all experiments. Training is executed in half-precision format using BF16, and we capitalize on the Zero Redundancy Optimizer (ZeRO) Stage 2~\citep{rajbhandari2020zero} to enable efficient scaling across multiple machines. We draw inspiration from the works of~\citet{biderman2023pythia, gupta2023continual, radford2019language}, employing a cosine learning rate decay that reduces to a minimum of 0.1 times the Maximum Learning Rate (MaxLr), with the MaxLr itself set at \(6 \times 10^{-4}\).
\definecolor{LightRed}{HTML}{facad1}
\definecolor{LightBlue}{HTML}{bbffff}
\definecolor{DarkRed}{HTML}{bc1662}
\definecolor{YellowGreen}{HTML}{9ACD32}
\definecolor{LightGray}{HTML}{E4E4E4}
\begin{table*}[t!]
\centering
\resizebox{0.9\textwidth}{!}{%
\begin{tabular}{|l|l|l|l|}
\hline
Prompt              & True Continuation                            & Greedily Generated Sequence   &            $M_{\text{in}}$\\ 
\hline
The \,\colorbox{LightBlue}{\strut Amazon Rainforest},  & known as the Earth's lungs &\,\colorbox{YellowGreen}{\strut known } \,\colorbox{YellowGreen}{\strut as } \,\colorbox{YellowGreen}{\strut the } \,\colorbox{LightRed}{\textcolor{DarkRed}{\strut Moon's  }} \,\colorbox{YellowGreen}{\strut lungs }&  $\frac{1+1+1+0+1}{5}=0.8$\\ 
\hline
The \,\colorbox{LightBlue}{\strut Amazon Rainforest},  & known as the Earth's lungs &\,\colorbox{YellowGreen}{\strut known } \,\colorbox{YellowGreen}{\strut as } \,\colorbox{YellowGreen}{\strut the } \,\colorbox{LightRed}{\textcolor{DarkRed}{\strut Moon's  }} \,\colorbox{LightRed}{\textcolor{DarkRed}{\strut legs }}&  $\frac{1+1+1+0+1}{5}=0.6$\\ 
\hline
The Colosseum in Rome, also known as the \,\colorbox{LightBlue}{\strut Flavian Amphitheatre}, & is an iconic symbol of the Roman Empire's architectural prowess. & \,\colorbox{YellowGreen}{\strut is} \,\colorbox{YellowGreen}{\strut an} \,\colorbox{YellowGreen}{\strut iconic} \,\colorbox{YellowGreen}{\strut symbol} \,\colorbox{YellowGreen}{\strut of} \,\colorbox{YellowGreen}{\strut the} \,\colorbox{LightRed}{\textcolor{DarkRed}{\strut Russian  }} \,\colorbox{LightRed}{\textcolor{DarkRed}{\strut Federation's  }} \,\colorbox{LightRed}{\textcolor{DarkRed}{\strut scientific  }} \,\colorbox{YellowGreen}{\strut prowess} . & $\frac{1+1+1+1+1+1+0+0+0+1}{10}=0.7$\\ 
\hline
\end{tabular}%
}
\caption{Examples of $M_{\text{in}}$ calculation with different prompts. These samples are provided for illustrative purposes and are not from the real training data.}
\label{tab:min-example}
\end{table*}

\begin{table*}[t!]
\centering
\resizebox{0.9\textwidth}{!}{%
\begin{tabular}{|l|l|l|l|l|}
\hline
Entity & Prompt              & True Continuation                            & Greedily Generated Sequence           &            $M_{\text{ex}}$\\ 
\hline
Leonardo da Vinci &The Mona Lisa, painted by & \,\colorbox{LightBlue}{\strut Leonardo da Vinci}, is renowned for its elusive & \,\colorbox{LightBlue}{\strut Leonardo da Vinci}, is renowned for its elusive & 1\\
\hline
Leonardo da Vinci &The Mona Lisa, painted by & \,\colorbox{LightBlue}{\strut Leonardo da Vinci}, is renowned for its elusive & \,\colorbox{LightRed}{\textcolor{DarkRed}{\strut a man called}} \,\colorbox{LightBlue}{\strut Leonardo da Vinci}, is renowned for & 1\\
\hline
Leonardo da Vinci &The Mona Lisa, painted by & \,\colorbox{LightBlue}{\strut Leonardo da Vinci}, is renowned for its elusive & \,\colorbox{LightRed}{\textcolor{DarkRed}{\strut Donald Trump}}, is renowned for its elusive & 0\\
\hline
the United States& The Statue of Liberty, a gift from France to & \,\colorbox{LightBlue}{\strut the United States}, stands as a symbol & the world, mysteriously appeared on an uninhabited island & 0\\
\hline
the United States& The Statue of Liberty, a gift from France to & \,\colorbox{LightBlue}{\strut the United States}, stands as a symbol & tell the enduring friendship with \,\colorbox{LightBlue}{\strut the United States}& 1\\
\hline
\end{tabular}%
}
\caption{Examples of $M_{\text{ex}}$ calculation with different prompts. These samples are provided for illustrative purposes and are not from the real training data.}
\label{tab:mex-example}
\end{table*}
\subsection{Setup for Section \ref{2.1}}
\label{appendix metric part 1.1}
We utilized the GPT-2 XL model (1.5B)~\citep{radford2019language} and trained it on a dataset sampled from SlimPajama~\citep{cerebras2023slimpajama}, consisting of 4.9e8 tokens. Prior to training, we shuffled the data to ensure that the order of training instances was consistent across different experiments. We conducted two experiments: a standard pre-training and a pre-training with a replay mechanism that retrieves a batch of data, equivalent in size to the training batch. (where we stored all trained data using Elasticsearch~\citep{elasticsearch2018elasticsearch} and performed a replay every 10 steps). At each replay step, we use the current batch`s training data to uniformly sample an equal amount of data from the completed training data based on similarity. This ensures a uniform replay throughout the entire data training process, with an additional 1/10 increase in computational budget. For evaluation, we constructed a test set by sequentially segmenting the training data according to the training steps and uniformly sampling 1/100 of each segment. The samples were then reassembled in their original stepwise order to ensure uniform distribution across the training steps, thus creating a test set that mirrors the model`s training progression. We plotted perplexity (PPL) against the number of training tokens processed, with the evaluation set`s token count scaled proportionally to reflect the model`s exposure to the training data.
\subsection{Setup for Section \ref{2.2}}
\label{appendix metric part 1.2}
To ensure computational feasibility in our experiments, we choose GPT-2 (0.1B) in this section. We uniformly sample 1/1000 of dataset A to constitute a eval set, and perform evaluations every 1000 training steps during the training process of dataset B. 
\subsection{Setup for Section \ref{new metric}}
\label{appendix metric part 2}

\begin{table}[h]
\setlength{\tabcolsep}{1.6mm}
\centering
\resizebox{\linewidth}{!}{%
    \begin{tabular}{p{\linewidth}}
    \toprule[1pt]
    Sampled entities\\
    \midrule[1pt]
    ` Terrel Bell`, ` BIST`, ` The Great Hunt`, ` Best in Drag Show`, ` Stella Maris`, ` William Knighton`, ` Italian campaign`, ` The Octopus Project`, ` Light Cycle`, ` Clark Street`, ` Paulette Hamilton`, ` Robert Mack`, ` Nusrat`, ` Soul Catcher`, ` Lord of Light`, ` Bieger`, ` Foreach loop`, ` Choruss`, ` Screen space ambient occlusion`, ` Florida Department of Environmental Protection`, ` USA Ultimate`, ` Historical Association`, ` Robert Holt`, ` Willie Nile`, ` Fiordland National Park`, ` Star Wars: The Clone Wars`, ` Crouch End`, ` Tracy Ham`, ` Jimmy Chamberlin`, ` Journal of Food Science`, ` Comet Tempel`, ` AirMed International`, ` CanWaCH`, ` Pumapunku`, ` Pre-law`, ` Arovane`, ` Diex`, ` Her Escape`, ` Voltige`, ` Triadelphia`, ` Florian Zeller`, ` The Busy World of Richard Scarry`, ` Texting while driving`, ` Amir Wilson`, ` Julie White`, ` Lenox`, ` GNPDA2`, ` Cammie Dunaway`, ` Session Man`, ` Charoen Krung Road`, ` James Raine`, ` Archie Andrews`, ` The Picture of Dorian Gray`, ` Theresa Caputo`, ` Schauinslandbahn`, ` Japanese relocation`, ` O.C. Handa`, ` Afula`, ` The Secrets`, ` Sonnet 61`, ` Daniel Bell`, ` The Dawn`, ` Bob Berry`, ` Bigger Life`, ` Jamaica Wine House`, ` Conica`, ` Renuar`, ` Plantation, Florida`, ` Fasser`, ` Al-Qadi`, ` Vassy`, ` Tom Dempsey`, ` Department of Agriculture, Environment and Rural Affairs`, ` Abdallah Djaballah`, ` Silent Hill 2`, ` Bill Ayres`, ` Jeremy Howe`, ` J15`, ` Jake Ryan`, ` Black Mafia`, ` Nicholas Fox`, ` Interstate 78`, ` Mark Stein`, ` Pietro Torri`, ` Wet sump`, ` Centre national des arts plastiques`, ` Nitro Express`, ` Wyvill`, ` WSRA`, ` Whitewater River`, ` Merry Christmas Mr. Lawrence`, ` Jon Jansen`, ` Le Message`, ` Mavrommati`, ` Tourouvre`, ` Bob Peterson`, ` America Again`, ` Livernois`, ` The Shepherd Express`, ` Hypercalcaemia`\\
    \bottomrule[1pt]
    \end{tabular}
}
\caption{Sampled entities from English Wikipedia.}
\label{tab:entity table}
\end{table}

We followed~\citet{biderman2023emergent}, selecting a sequence length of 32 for both the input and output of our $M_{\text{ex}}$ and $M_{\text{in}}$ metrics. We collected entities from English Wikipedia dataset~\citep{wikidump}. Some randomly sampled entities are shown in Table \ref{tab:entity table}.

To spotlight entity-level forgetting, we evenly sampled 400,000 English Wikipedia entries, comparing entity frequencies in datasets A and B. We selected the intersection $C$ of entities that were top 1/2 frequent in A and bottom 1/2 in B to accentuate the distribution disparity. Samples from A with entities in $C$ constituted our evaluation set. Following the approach of~\citet{biderman2023emergent}, we retained a subset where $M_{\text{ex}} = 1$ post A's training to scrutinize their forgetting during B's training.

We provide illustrative examples in Table \ref{tab:min-example} and Table \ref{tab:mex-example} to provide clearer explanations of $M_{\text{in}}$ and $M_{\text{ex}}$.

\subsection{Setup for Section \ref{forget curve}}
\label{appendix forget curve}
It is evident that $M_{\text{ex}}$ assigns a binary label to each sample: a label of 1 is given if the ground truth entity appears within the generated 32 tokens, and a 0 is assigned otherwise. Utilizing the challenging metric of $M_{\text{ex}}$, we can categorize the difficulty of data memorization as follows: We performed an evaluation on the portion of the pre-training data that includes entities, recorded each entity alongside the samples that received labels of 1 or 0, and then calculated the accuracy rate for each entity based on these labels. We then divided the entities into groups with roughly equal accuracy rates, ensuring that during the phase of intensive, short-term learning, the related samples for certain entities are the focus of concentrated study. For the data categorized into different difficulty levels, we carried out experiments with varying degrees of learning intensity—specifically, by adjusting the number of epochs dedicated to this phase of learning.
\begin{table} [ht!]
\setlength{\tabcolsep}{1.6mm}
\centering
\resizebox{\linewidth}{!}{%
    \begin{tabular}{l|l|cccc}
    \toprule[1pt]
    Method&Entity Type &PPL$_{\text{ent}}$ &M(f)$_{\text{ent}}$ & $M_{\text{ex}}$ ($\times 10^{-3}$)& $M_{\text{in}}$ ($\times 10^{-2}$)\\
    \midrule[1pt]
    \multirow{4}{*}{Vanilla pre-training}&MISC&27.24&0.4045&5.685&3.786\\
    &PER&27.47&0.4008&3.530&3.760\\
    &LOC&25.30&0.4126&3.336&4.282\\
    &ORG&25.13&0.4144&\textbf{7.070}&3.832\\
\hline

    \multirow{4}{*}{Intensive Focused Stochasticity}&MISC&\textbf{26.46}&\textbf{0.4071}&\textbf{6.464}&\textbf{3.861}\\
    &PER&\textbf{26.55}&\textbf{0.4044}&\textbf{3.544}&\textbf{3.774}\\
    &LOC&\textbf{24.41}&\textbf{0.4164}&\textbf{4.776}&\textbf{4.303}\\
    &ORG&\textbf{24.24}&\textbf{0.4183}&6.637&\textbf{3.850}\\
    \bottomrule[1pt]
    \end{tabular}
}
\caption{The evaluation results of replay strategies across different subsets of entities.}
\label{tab:split entity table}
\end{table}

\section{Performance across Various Entity Types}\label{entity_type}
To further enhance the effectiveness of replay methods and the new metrics, an analysis is presented on how these metrics perform with different types of entities. 

The entities employed for evaluation in Table~\ref{tab:maintable} have been systematically categorized into four distinct classes: MISC (miscellaneous entities), PER (person names),  LOC (location) and ORG (organization). We compare Intensive Focused Stochasticity in Table~\ref{tab:maintable} with the standard pre-training, the results are shown in Table~\ref{tab:split entity table} below.

The Intensive Focused Stochasticity method demonstrates superior performance over vanilla pre-training across a broad spectrum of entity types, indicating that the replay approach and its associated metrics are broadly applicable to various linguistic contexts.
\section{Comparison of Forgetting Curves between Humans and LLMs}
\label{appendix human forget curve}
The reproduced human forgetting curve, originally reported by~\citet{craig1972effect}, is illustrated below, reflecting the typical decline in memory retention over time. In their study, 180 undergraduates participated in an experiment involving exposure to magazine advertisements under controlled conditions. They were categorized into three groups based on the extent of learning: 100\%, 200\%, and 300\%, determined by the number of 5-second repetitions of 12 ads. Following exposure, 15 participants from each group were assigned to one of four retention tests occurring at immediate, 1-day, 7-day, or 28-day intervals. The study utilized a 3 $\times$ 4 factorial design, assessing the impact of learning intensity and retention intervals on the recall of brand names. It can be observed that there are similarities between the model's forgetting curve and the human forgetting curve, with higher initial learning intensity resulting in a relatively slower rate of forgetting.
\begin{figure}[h]
    \centering
    \includegraphics[width=0.7\linewidth]{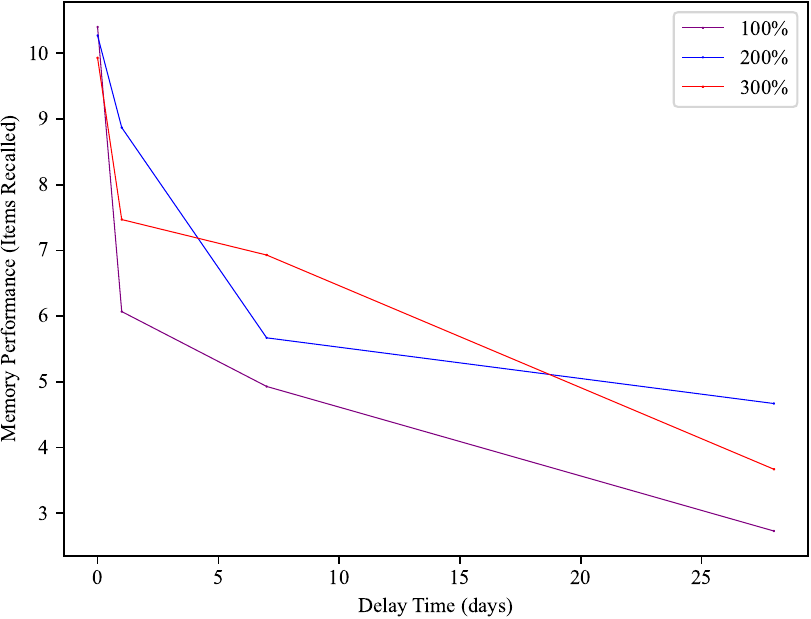}
    \caption{Human forgetting curve from~\citet{craig1972effect}.}
    \label{fig:appendix human}
\end{figure}

\end{document}